\documentclass[preprint, review, 3p, authoryear, times]{elsarticle}

\makeatletter
\def\ps@pprintTitle{%
 \let\@oddhead\@empty
 \let\@evenhead\@empty
 \let\@oddfoot\@empty
 \let\@evenfoot\@empty
}
\makeatother

\usepackage{graphicx}
\usepackage{url}
\usepackage{color, soul}
\usepackage{bm}
\usepackage{amsmath} 
\usepackage{lineno}

\usepackage{color}

\usepackage{array}
\usepackage{graphicx}
\usepackage{geometry}
\usepackage{listings}
\usepackage{xcolor}
\usepackage{hyperref}
\usepackage{svg} 

\usepackage{subcaption} 
\newcommand{\rev}[1]{\textcolor{black}{#1}}
\newcommand{\revv}[1]{\textcolor{black}{#1}}

\usepackage{graphicx} 
\usepackage{listings} 
\usepackage{array} 

\usepackage{verbatim}

\usepackage{graphicx}
\usepackage{caption}
\usepackage{subcaption}
\usepackage{pdflscape}

\begin{document}

 \title{   
    Towards Next-Generation Urban Decision Support Systems through AI-Powered \revv{Construction} of \rev{Scientific Ontology} using Large Language Models - A Case in Optimizing Intermodal Freight Transportation
 } 


 \author[UTK]{Jose Tupayachi}\ead{jtupayac@vols.utk.edu}
 \author[CSED]{Haowen Xu \corref{cor1}} \ead{xuh4@ornl.gov}
 \author[CSED]{Olufemi A. Omitaomu}\ead{omitaomuoa@ornl.gov}
 \author[UTK]{Mustafa Can Camur}\ead{mcamur@utk.edu}
 \author[UTK]{Aliza Sharmin}\ead{asharmin@vols.utk.edu}
 \author[UTK]{Xueping Li}\ead{xueping.li@utk.edu}

 \cortext[cor1]{Corresponding author.}

 \address[CSED]{Computational Urban Sciences Group, Oak Ridge National Laboratory, Oak Ridge, TN 37830, USA}
 \address[UTK]{Department of Industrial and Systems Engineering, The University of Tennessee, Knoxville, Knoxville, TN 37996, USA}

\makeatletter
\newcommand{\printfnsymbol}[1]{%
  \textsuperscript{\@fnsymbol{#1}}%
}

\newcommand*{\MyIndent}{\hspace*{0.5cm}}%


\begin{abstract} \label{sec:abstract}
The incorporation of Artificial Intelligence (AI) models into various optimization systems is on the rise. However, addressing complex urban and environmental management challenges often demands deep expertise in domain science and informatics. This expertise is essential for deriving data and simulation-driven insights that support informed decision-making. In this context, we investigate the potential of leveraging the pre-trained Large Language Models (LLMs) \rev{to create knowledge representations for supporting operations research}.
By adopting ChatGPT\revv{-4} API as the reasoning core, we outline an \rev{applied} workflow that encompasses natural language processing, methontology-based prompt tuning, and \rev{Generative Pre-trained Transformer (GPT)}, to automate the \revv{construction} of scenario-based ontologies using existing research articles and technical manuals of urban datasets and simulations. \rev{From these ontologies,} knowledge graphs \rev{can be derived} using widely adopted \revv{formats and protocols, guiding various tasks towards data-informed decision support.} The performance of our methodology is evaluated through a comparative analysis that contrasts our AI-generated ontology with the widely recognized Pizza Ontology, commonly used in tutorials for popular ontology software. We conclude with a real-world case study on optimizing the complex system of multi-modal freight transportation. \revv{Our approach advances urban decision support systems by enhancing data and metadata modeling, improving data integration and simulation coupling, and guiding the development of decision support strategies and essential software components.} 


\end{abstract}

\maketitle
\section{Introduction} 
The global urban population has surged from 1.01 billion in 1960 to 4.52 billion in 2022 and is expected to reach 6.9 billion, or approximately 70\% of the global population, by 2050 \citep{hoornweg2017population}. The accelerated pace of global urbanization underscores the urgent need for effective management and optimization of urban systems. The goals are to promote urban livability through efficient resource allocation \citep{habitat2020value} and reduce the global carbon footprint, thereby fostering sustainable economic growth amid increasing climate change pressures \citep{moran2018carbon}. Over recent decades, academia, industry, and governmental agencies have significantly invested in developing smart city applications \citep{angelidou2018enhancing}. These efforts aim to foster smart and sustainable cities to enhance urban planning and management, optimizing city operations through the integration of cutting-edge computing and communication technologies \citep{bibri2017core}. The synergy of these emerging technologies, specifically the Internet of Things (IoT), digital twins, and artificial intelligence (AI), alongside advanced simulations across various urban domains (including transportation, water distribution, buildings, and energy systems), promotes a more informed, reliable, and time-sensitive approach to complex urban decision-making \citep{rane2023integrating, xu2023smart}. The approach leverages large volumes of urban data and simulation outputs, combined with expert knowledge and experiences, to produce holistic and resilient strategies that enhance the efficiency of urban management and city operations \citep{xu2023toward, weil2023systemic}. 

Despite the successes of past smart city applications, challenges persist in optimizing complex urban systems, which involve intricate interactions among various subsystems \citep{ mcphearson2016advancing}. A prime example is the intermodal freight transportation system, spanning vast areas and governed by dynamic, multi-scale processes with broad socioeconomic and environmental impacts \citep{heinold2020emission}. \rev{Developing holistic solutions to optimize such a system often requires in-depth domain expertise and laborious efforts to conceptualize the problem, formulate decision support \revv{strategies}, identify relevant data sources \revv{and variables} that represent essential processes and dynamics, and create a comprehensive data provisioning system (e.g., database and repository) through the integration of diverse urban data and coupled simulation outputs across multiple domains. This data provisioning system is crucial for providing information to analytical and optimization models, as well as decision support workflows, to generate both model and data-driven insights essential for informed decision-making in urban digital twins and decision support systems.}

\rev{In this regard, many previous studies have taken a knowledge engineering approach, focusing on \revv{constructing} scenario-based ontologies to conceptualize complex processes in decision support problems \citep{matei2021multimodal}. They create knowledge graphs from these ontologies to support data integration and simulation coupling, thereby \revv{guiding the development of efficient} data provisioning systems \citep{chen2018ontology}. This ontology-driven approach is vital for improving problem understanding, \revv{identifying} objectives, and designing optimization workflows \citep{wang2019ontology, kornyshova2010decision}. However, this approach} is often constrained by the \rev{domain-specialized} nature of the work and the extensive interdisciplinary knowledge required to \revv{construct} scenario-based ontologies for addressing specific urban challenges.
\rev{In this regard, there is a need to develop an intelligent approach that leverages cutting-edge AI techniques to automatically collect, compile, and \rev{interpret} interdisciplinary domain knowledge from \revv{text documents}. This approach \revv{should} generate scenario-based knowledge representations, such as ontologies and knowledge graphs, to facilitate the development of \revv{decision support strategies and} data provisioning systems for optimizing complex urban systems.}

\begin{figure*} 
    \centering    
    \includegraphics[width=\textwidth]{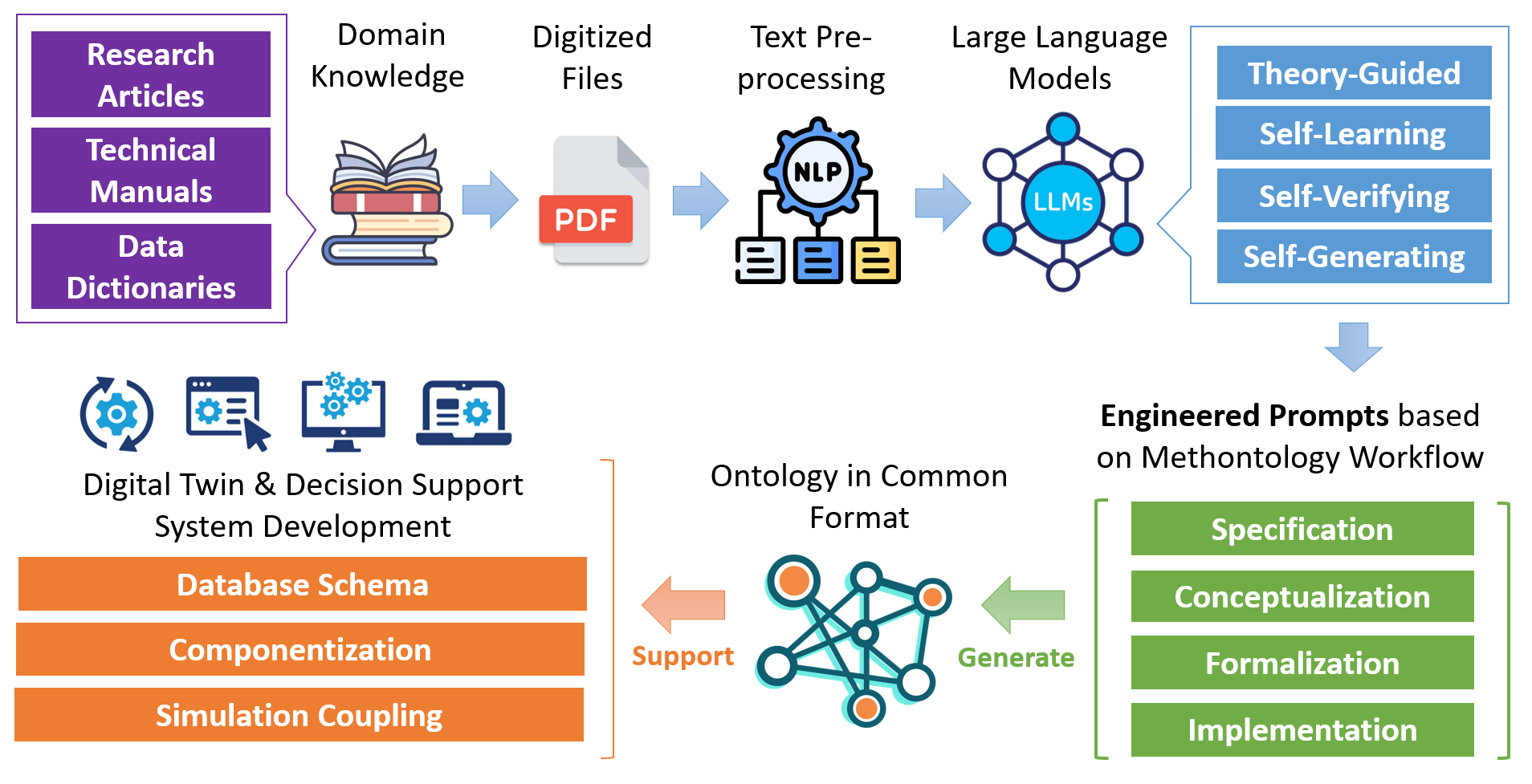}
    \caption{Conceptual design of an LLM-powered autonomous agent for automated ontology generation.}\label{fig:Infographic1}    
\end{figure*}

Recent progress in generative models, notably large language models (LLMs) like GPT-3 \citep{brown2020language}, GPT-4 \citep{achiam2023gpt}, and Google Gemini \citep{team2023gemini}, has significantly propelled the research and development of \revv{AI agents} and autonomous systems forward. In light of these developments, this paper explores the potential of LLMs to leverage their advanced understanding, reasoning, and generative abilities in human language \revv{to generate knowledge representations for operations research}. By employing the \revv{ChatGPT-4} API as a core reasoning tool, we outline a detailed workflow that incorporates (a) scholarly databases, (b) natural language processing tools, (c) structured prompts from the Methontology workflow, and (d) pre-trained LLM. \revv{The workflow is designed} to automate the \revv{construction} of scenario-based ontologies through the analysis of research articles and technical documents on \rev{recent emerging} urban data and simulations. \rev{The AI-generated ontologies expand the domain ontology by adding new data sources, \revv{variables, and simulation capabilities}. From these ontologies, we can drive} knowledge graphs in standard \rev{formats}, such as RDF, \rev{using knowledge engineering software and tool}. \rev{The derived knowledge graphs enhance various} aspects of urban decision support systems \rev{by enhancing} data \rev{and metadata} modeling, dataset integration, simulation \rev{coupling}, and the formulation of decision support \revv{strategies}. Further, we evaluate our method's effectiveness by comparing \rev{its ontology output} with the Pizza Ontology, \rev{which is often} used \rev{as the sample data} in tutorials \rev{of ontology software,} such as Protégé. The paper culminates in a case study on optimizing intermodal freight transportation in urban environments, showcasing how our \revv{AI-generated} ontologies \rev{and knowledge graphs are used to guide the integration, management, retrieval, \revv{and interpretation} of complex \revv{multi-dimension} datasets, as well as formulation of decision support strategies to} enable informed decision optimization.



\section{Literature Review} \label{sec:literature_review}
Ontologies and knowledge graphs are foundational elements in the realms of the semantic web and knowledge management, providing structured ways to represent knowledge about the world. Defined as ``a formal, explicit specification of a shared conceptualisation" by  \cite{noy1997state}, ontology, is a formal representation of knowledge as a set of concepts within a domain, and the relationships between those concepts \citep{gruber1993ontology}. It also provides standard vocabularies for researchers to annotate, share, and reuse their knowledge across different disciplines \citep{chen2018ontology}. Knowledge graphs further extend this concept by creating large networks of entities and their interrelations, encapsulated in graph structures, which enable complex queries and analytics, making knowledge more accessible and actionable \citep{ji2021survey}. 

\subsection{\rev{Knowledge Representation} for Smart Cities} \label{subsec:ONtology_definition}
In smart city applications, ontology and knowledge graphs play pivotal roles in enhancing decision support and optimizing urban system. They facilitate the integration of diverse data sources, provide a basis for interoperability between different systems and platforms, and support the reasoning processes needed for intelligent decision-making \citep{chen2018ontology}. By mapping the complex relationships between various urban elements, such as transportation networks, utilities, and services, these knowledge representations synthesize data into actionable insights, thereby improving efficiency, sustainability, and the quality of urban life. \citep{bibri2017core}. 

\revv{From the technical perspective}, ontology and knowledge graphs are crucial in the development of web-based decision support systems \citep{spoladore2021collaborative}, digital twins \citep{petrova2021digital}, databases \citep{balasubramani2016ontology}, AI applications \citep{bergman2018knowledge}, and software dedicated to \revv{promote} smart city management \citep{shi2023ontology}. \rev{Ontologies, with their hierarchical knowledge representations, provide the underlying architecture to structure and connect these technologies into meaningful software components and workflows, dynamically modeling and simulating the real-world urban environment \citep{al2019coupling}. This configuration enables cities to anticipate problems, optimize operations, and engage more effectively with citizens \citep{kuster2020udsa}. Knowledge graphs, derived from ontologies, are topological networks that represent real-world entities (e.g., objects, events, situations, or concepts) and illustrate the relationships between them \citep{simsek2022knowledge}. These networks are more application-focused in the context of data integration, management, and retrieval systems. They are widely applied in urban data applications to facilitate semantic search, recommendation systems, and AI models to link and contextualize disparate data \citep{chaudhri2022knowledge, syed2022context}.} Through the use of these semantic technologies, smart city applications, such as digital twin and online decision support systems, can evolve into more adaptive, responsive, and intelligent software applications, harnessing the full potential of the online big urban data to meet the challenges of complex urban system management and optimization \citep{weil2023urban}. 


\subsection{Ontology \revv{Construction}} \label{Ontology Design and Development} 
Methodologies for constructing scientific ontologies often incorporate both detailed and systematic approaches to effectively address the multifaceted nature of ontology development \citep{cristani2005survey}. At the micro level, these methodologies are crucial for focusing on the intricate details of ontology creation, such as the precise formalization of axioms and the design decisions shaped by the selected representation language \citep{guizzardi2005ontological}. This attention to detail ensures the accuracy and consistency of the ontology, which are vital for its reliability and usability. Conversely, macro-level approaches are essential for providing a holistic systems perspective, allowing for a comprehensive understanding of the development process through the lens of information systems and IT \citep{merali2006complexity}. By considering the broader context, these approaches facilitate the integration and alignment of the ontology with existing systems and processes, ensuring its scalability and adaptability. Together, these methodologies offer a balanced and thorough approach to ontology development, accommodating both specific technical considerations and overarching structural coherence. Well-known methodologies for constructing ontologies include older frameworks like Methontology and On-To-Knowledge, alongside more contemporary ones that focus on the re-usability and flexibility of ontology, such as NeON.
\begin{itemize}
    \item Methontology is a rigorous framework developed for the creation of domain ontologies that are independent of applications, facilitating a structured approach to ontology design \citep{gosal2015ontology}. This methodology encompasses several key phases, including specification, conceptualization, formalization, implementation, and maintenance, ensuring a comprehensive development process. Originally applied in the late 1990s, Methontology emphasizes systematic steps for the effective organization and representation of knowledge, making it a cornerstone in the field of semantic technologies and knowledge engineering \citep{fernandez1997methontology}. 
    \item The On-To-Knowledge methodology (OTKM) represents a structured approach to ontology design, aimed at enhancing knowledge management within organizations. It focuses on converting implicit company knowledge into an explicit, structured form, facilitating better decision-making and process optimization \citep{sure2004knowledge}. This method emphasizes early stakeholder involvement, iterative development, and the use of formal ontologies to ensure that the captured knowledge is both accurate and usable in automated systems.
    \item The NeON methodology caters to the evolving demands of ontology engineering, supporting the reuse, re-engineering, and integration of ontological resources \citep{gomez2009neon}. By offering a diverse set of strategies and tools, NeON facilitates the development of robust ontologies and ontology networks, addressing complex semantic frameworks \citep{suarez2015neon}. This methodology is particularly beneficial for projects requiring collaborative and distributed development, making it a pivotal resource for advancing semantic technologies and their applications in various domains.    
\end{itemize}

The ontology development landscape has evolved towards a collaborative and distributed paradigm, transitioning from the creation of singular, monolithic ontologies to the collaborative construction of ontology networks \citep{keet2004aspects}. Consequently, contemporary methodologies adapt to the nuances of collaborative and distributed development, supporting concurrent contributions from domain experts and knowledge engineers across different locales and encouraging a modular approach for enhanced management and scalability \citep{nguyen2011ontologies}. Each methodology contributes a unique perspective to ontology development, focusing on intricate authoring techniques or overarching system-wide processes. The selection of a methodology can profoundly affect the development trajectory, influenced by variables such as the application domain, team composition, and the intended utility of the ontology.

\subsection{Traditional Methods for Ontology Implementation} \label{sec:tradMethod}
At the implementation level, a variety of software tools for ontology creation, visualization, manipulation, and editing are widely utilized by knowledge engineers and domain scientists alike. Among these, Protégé \citep{noy2001creating}, OntoSoft \citep{gil2016ontosoft}, Hozo \citep{mizoguchi2007model}, and OntoGraf \citep{falconer2010ontograf}, yEd Graph Editor \citep{sedlmeier2016model}, Visual Understanding Environment (VUE) \citep{kumar2008visual}, and CmapTools \citep{canas2004cmaptools} have emerged as notable options in recent years. When choosing these tools, ontology developers must consider several critical factors, including the capability for reusing existing ontologies, comprehensive documentation, as well as the ability to export and import data across diverse formats, views, and libraries \citep{dudavs2018ontology}. These considerations are essential for ensuring the effective application and integration of ontological tools in scientific and engineering workflows.

Implementing ontology with specialized management software is a time-consuming and complex process, largely due to the detailed work required to represent domain-specific knowledge accurately and the expertise needed to effectively utilize these tools \citep{botzenhardt2011developing}. The development of ontology is inherently iterative, involving repeated cycles of testing, evaluation, and refinement. Each cycle can reveal new requirements or inconsistencies, necessitating a return to previous stages, which extends the duration of the project \citep{fernandez2002overview}. Given these challenges, there is a pressing need to harness emerging AI technologies to automate the ontology \revv{construction}, potentially streamlining the process and reducing the manual effort involved.


\subsection {Semantic Web Standards} \label{Semantic Web Standards}
After being implemented, scientific \revv{knowledge representations} are typically stored and documented using common semantic web languages that adhere to the semantic web standards, including RDF (Resource Description Framework), OWL (Web Ontology Language), and SPARQL (SPARQL Protocol and RDF Query Language) \citep{allemang2011semantic}. These standards are crucial for structuring, storing, and querying data effectively. They enable the structured representation and interconnected querying of data across diverse systems, facilitating seamless data integration and accessibility on the Semantic Web.
\begin{itemize}
    \item RDF serves as a standard model for data interchange on the web, enabling the representation of information in a triple format (subject-predicate-object), facilitating a graph-like structure \citep{tomaszuk2020rdf}.
    \item OWL provides additional vocabulary along with formal semantics designed for processing and integrating information as it enables the creation of more complex ontologies \citep{cardoso2015web}. 
    \item SPARQL, on the other hand, is used to query databases stored in RDF format. It allows for powerful and expressive queries over diverse data sources, making it essential for extracting and manipulating data within and across knowledge graphs \citep{cure2014rdf}. 
\end{itemize}
 Together, these semantic web languages form the backbone of semantic web applications, allowing for robust and sophisticated handling of scientific resources, such as the sharing of scientific datasets and models, across the internet. Many smart city applications depend on semantic web technologies to discover and integrate essential data for \revv{promoting a variety of smart city services}. Consequently, these technologies play a pivotal role in the development of numerous urban decision support systems \revv{and digital twins}. Therefore, it is imperative that future AI-powered techniques for constructing ontologies adhere to established Semantic Web Standards. This alignment will ensure compatibility and enhance the effectiveness of data integration across diverse platforms and systems. 

\subsection {AI-powered Autonomous System}
\label{subsec:ai-powered}
The swift evolution of AI technologies has catalyzed the emergence of autonomous systems\citep{brustoloni1991autonomous}, which are sophisticated computer systems designed to undertake tasks and make decisions autonomously, with little to no human oversight \citep{wooldridge1995intelligent}. These agents have demonstrated significant potential across diverse fields such as robotics, healthcare, transportation, and finance, showcasing benefits like enhanced efficiency, diminished errors, and the capability to manage intricate tasks that might be impractical or exceedingly laborious for humans \citep{ stone2022artificial}. Diverging from automated systems, such as automated irrigation systems and home automation systems, \revv{which} execute predefined tasks based on specific sets of commands, inputs, rules, and environmental conditions, autonomous systems exhibit higher complexity, better adaptability to their surroundings, and the ability to make informed decisions and content generation from collected data \citep{albrecht2018autonomous}. 

The capability of most existing rule- or algorithm-based reasoning cores is confined to certain or closed environments, limiting their reasoning capacity for tasks that involve creating new content or computer codes. With the advancement of generative AI models, LLMs, such as GPT-3 \citep{brown2020language} and GPT-4 \citep{sun2023gpt}, has significantly propelled forward the research and development of autonomous systems. These models exhibit a profound comprehension of human natural language, enabling them to adeptly perform tasks across various domains including reasoning, creative writing, code generation, translation, and information retrieval \citep{li2023autonomous}. With recent advancements in prompt engineering and tuning techniques, LLMs are increasingly being integrated as the reasoning core to enable AI-powered assistants capable of performing specialized tasks, such as generating content within specific domain contexts, automating code generation, and synthesizing data \citep{shin2023prompt}. This integration highlights the transformative impact of LLMs in enhancing automated processes and driving innovation across various fields, shedding light on an autonomous approach for generating scientific ontologies. 

\subsection{\revv{LLM-powered Knowledge Engineering}}
In recent years, there is an emerging trend to \revv{explore the integration} of Large Language Models (LLMs) and Retrieval-Augmented Generation (RAG) techniques with knowledge representations. This integration aims to enhance the accuracy and reliability of generative AI models, particularly when they are tasked with answering specialized scientific questions \citep{meyer2023llm, baldazzi2023fine}. \rev{As a majority of current studies focus on utilizing existing ontologies and knowledge graphs to improve the response accuracy of large language models (LLMs) and reduce potential hallucinations, there are only a \revv{few ongoing} studies, \revv{most of which are presented as arXiv preprints,} that have explored the potential of LLMs for constructing ontologies and knowledge graphs.}

\rev{
\cite{babaei2023llms4ol} introduces LLMs4OL, leveraging LLMs for Ontology Learning (OL). The study evaluates nine LLM models using zero-shot prompting for term typing, taxonomy discovery, and non-taxonomic relation extraction across domains like WordNet, GeoNames, and UMLS. Results show that while foundational LLMs alone may not suffice for complex ontology construction, fine-tuning can alleviate knowledge acquisition bottlenecks. \cite{zhang2024extract} presents the Extract-Define-Canonicalize (EDC) framework for automating knowledge graph creation using LLMs. EDC overcomes schema context limitations by splitting the process into open information extraction, schema definition, and canonicalization, handling both predefined and auto-constructed schemas. It outperforms state-of-the-art methods in extracting high-quality triplets across benchmark datasets. \cite{kommineni2024human} explores LLMs in automating ontology and knowledge graph construction using , introducing a semi-automated pipeline that minimizes human involvement while maintaining quality. This method, applied to deep learning methodologies, shows that LLMs can effectively assist with reduced human effort, though human validation remains crucial. \cite{caufield2024structured} describes SPIRES, a method using LLMs for zero-shot learning and recursive prompt interrogation to automate knowledge base and ontology construction. SPIRES ensures accurate grounding with existing ontologies and excels across various domains, offering flexibility and ease of customization. It supports combining human expertise with advanced AI techniques to convert unstructured text into structured knowledge and is available in the open-source OntoGPT package.
}

\section {Knowledge Gaps and Contributions}
\label{subsec:challange-gap}
Despite the availability of well-established methodologies, ontology management software, and advanced web semantic languages for designing, developing, and implementing scientific ontologies, \rev{enabling automated} creation of efficient and reusable ontologies to support smart city services often encounters \revv{challenges and technical limitations}.

\subsection{\revv{Domain Application Challenges}}
\revv{In the field of smart city management, creating scenario-based ontologies is crucial for enhancing data discovery, integration, and management across various urban sectors, such as parking availability, building energy optimization, and traffic volume prediction. This is a sophisticated task that often requires large volumes and diverse types of data, as well as outputs from simulations and predictive analytics. The efforts involved in data preparation and provisioning are also essential for formulating informed decision support. However, these tasks are typically repetitive, time-consuming, and demand specialized domain expertise.}

\revv{As new data sources and urban simulation scenarios rapidly emerge across expansive urban regions, and with the deployment of new sensors and IoT-connected infrastructure, the challenge intensifies. Additionally, the increasing availability of data from various sources, such as crowd-sourcing, autonomous vehicles, and simulation outputs, further complicates the landscape. From a data interoperability perspective, ontologies and knowledge graphs play critical roles in multi-domain data integration and simulation coupling \citep{xu2020web, chen2018ontology}. For instance, solving complex optimization problems in freight transportation systems often requires integrating data on freight, fuel and emission simulations, real-time traffic, and transportation infrastructure. These comprehensive tasks can definitely benefit from an ontological approach. Developing system-based knowledge representation using ontologies and knowledge graphs is essential for discovering, managing, and sharing complex urban science data, which are indispensable for creating urban digital twins, information systems, and decision support tools.}

\revv{Despite the critical importance of urban knowledge representation and semantic techniques in the smart city sector, designing and constructing scenario-based ontologies for newly emerging data sources or simulation models often require significant time and specialized effort. This process demands deep domain expertise to extract and interpret essential knowledge from extensive documentation, research articles, and manuals. After extracting core knowledge and representing it as high-level concepts, properties, and relationships, it is vital to maintain the semantic consistency of the derived ontology with existing domain ontologies. This ensures that new data and simulations are interpreted correctly, uniformly, and can be seamlessly integrated into existing \revv{data and} software tools. However, this task involves complex and iterative processes, including data and metadata standardization, semantic mapping, and thorough validation and testing. Consequently, there is a pressing need for \revv{automated} methods to design and construct ontologies and knowledge graphs that \revv{describe} newly emerging urban data and simulation outputs.}


\subsection{\revv{Methodological Challenges}}
\revv{There has been a recent surge in research interest in developing more efficient and automated methods for ontology and knowledge construction by leveraging cutting-edge LLMs. However, existing studies in this area are still rare and in their preliminary stages. Our review identified only four relevant studies, including a few ongoing research efforts presented through preprints.}

\revv{Among these, two contemporary studies focus on proving the feasibility of using LLMs to construct ontologies and knowledge representations from text inputs using zero-shot prompting methods and general-purpose queries \citep{babaei2023llms4ol, caufield2024structured}. These studies provide important proof-of-concepts, demonstrating the potential of LLMs with simple text documents. However, they do not focus on training or guiding LLMs with standard, system-based knowledge engineering methodologies to tackle complex ontology construction tasks involving sophisticated scientific data, which encompasses quantitative fields and multiple data features (such as spatial, temporal, and topological dimensions). The proper interpretation of logic and entity relationships among multiple quantitative fields in scientific data and simulation outputs has not been addressed in previous studies, often requiring LLMs to follow specialized knowledge engineering methodologies. One study explored the potential for automating the creation of domain ontologies and knowledge graphs using text inputs, LLMs, and RAG through the pre-definition of competency questions \citep{kommineni2024human}. This approach aims to formulate a more self-organizing method for creating foundational domain knowledge using diverse text content. However, the research does not focus on creating scenario-based ontologies and knowledge graphs based on specific data sources and simulation tools with particular application objectives for supporting operations science. Another study proposed a novel entity, domain, and context (EDC) method to improve the extraction of entity-relation triplets from generic real-world text documents and automate the generation of knowledge graphs \citep{zhang2024extract}. This study implemented its proposed method using multiple LLM frameworks and compared their performances. Despite its unique contributions, the study does not emphasize constructing ontological knowledge representation in a hierarchical structure but instead focuses on encoding knowledge into a graph structure. Additionally, the study is not designed to extract knowledge and information from domain-specific text inputs that describe scientific datasets and simulation tools. As a result, it does not emphasize domain-specific relation extraction, definition, or knowledge fusion with existing domain ontologies.}

\subsection{\revv{Knowledge Gaps}}
\revv{Based on our literature review, we have identified several gaps in the existing studies on LLM-ontology. The following aspects have been rarely addressed in previous research.}
\revv{
\begin{description}
    \item[Complex Domain applications in Operations Science:] Many previous studies have concentrated on testing the feasibility of constructing ontologies and knowledge using LLMs through proof-of-concept and pilot studies. However, there is a gap in utilizing AI-generated ontologies to support real-world decision-making and evaluating their performance in practical applications.     
    \item[Building Concepts with Complex Features:] 
    Most previous studies have utilized generic text inputs to test the ontology and knowledge graph creation capabilities of LLMs. However, these studies did not involve the handling and interpretation of complex scientific datasets with various data dimensions and features, such as topological, spatial, and temporal aspects. 
    \item[Integration of Knowledge Engineering Methods:] Many previous studies have employed general queries with zero-shot prompting methods to generate knowledge representation using LLMs. However, the aspect of training or instructing LLMs using commonly-accepted knowledge engineering methodologies, as discussed in Section \ref{sec:tradMethod}, has been rarely addressed.
    \item[Interoperability of LLM-generated Content:] While previous studies have focused on testing the feasibility and accuracy of LLM-generated knowledge representations, they often overlook the aspects of interoperability and usability, particularly concerning their application to practical research problems. Potential application areas include supporting the formulation of decision support strategies, guiding the design of critical software and database components, and facilitating the construction of knowledge bases for practical purposes. 
\end{description}
}
\revv{
Based on the findings and insights from previous studies, we believe there is a need to test the LLM-ontology capabilities on more complex scientific datasets, simulations, and their manuals. This will help create AI-powered methodologies that support knowledge extraction and management, ultimately addressing complex problems in operations science.}


\subsection{\revv{Motivation and Contributions}}
\revv{
Building on prior research into the use of Large Language Models (LLMs) for constructing ontologies and knowledge graphs, we aim to advance the current state-of-the-art by exploring LLMs' capabilities to identify, extract, and interpret concepts in more sophisticated scientific datasets and simulation outputs, which are often quantitative and multi-dimensional. Our goal is to create knowledge representations with practical benefits to support complex decision-making in operations research.}

\revv{Based on this motivation, we present the design, implementation, and application of an integrated methodology that utilizes the well-accepted METHONTOLOGY framework to instruct an LLM in a few-shot learning fashion. This approach aims to construct scenario-based ontologies from urban research articles, technical manuals, and data subsets. An integral part of our methodology includes a workflow that provides options for utilizing the AI-generated knowledge representations to (a) guide the design and development of critical software and database components for urban decision support systems and (b) facilitate the formulation of decision support strategies, which involve the definition decision metrics and workflows. Due to the interdisciplinary nature of our study, our research contributes to the fields of operations science, software engineering, and the applications of LLMs.}
\revv{
\begin{description}    
    \item[Contributions to LLM Application:] We explored the ability of Large Language Models (LLMs) to understand and interpret entities and relationships in complex spatial and topological datasets and simulation outputs, guided by the METHONTOLOGY framework. Our approach is demonstrated through Section \ref{sec:OKGC}.
    \item[Contributions to Software Engineering:] We demonstrate a practice of utilizing AI-generated ontology to guide software design and semi-automatically generate critical components of a data provisioning system essential for developing an urban decision support system. Our approach aims to systematically bridge AI-generated knowledge representation with existing ontology management tools and database technologies (detaied in Section \ref{sec:DPSDI}), supporting practical tasks in the development of decision support software.
    \item[Contributions to Operations Science:] We devised an automated AI-powered workflow, driven by integrated data, simulations, and knowledge representations, to develop intelligent and advanced capabilities for supporting decision-making and optimizing complex urban systems. A detailed demonstration is presented through Section \ref{sec:DSWF}. 
\end{description}
}

\section{Methodology} \label{sec:Method}
This section begins with an overview of the background and design requirements for our proposed ontology generation workflow. It then details the development of our methodology, both conceptually and technically. The ontology produced through our approach is validated by comparing it with the well-known Pizza Ontology, which is frequently used in tutorials for popular ontology software.

\subsection{Design Requirements} \label{sec:terminology}
Our initiative to prototype an LLM-powered workflow for ontology generation is driven by the need to create a large-scale, data-driven digital twin designed to support complex decision-making for optimizing the intermodal transportation system across the contiguous US. Our digital twin, debuted as ARPA-E RECOIL, leverages the latest state-of-the-art intermodal transportation datasets and simulation tools to optimize the freight transport system. This comprehensive approach takes into account supply and demand dynamics, delivery times, and costs, along with potential socioeconomic impacts such as traffic congestion, air pollution, and carbon emissions. Consequently, it necessitates the creation of an ontology behind the proposed urban digital twin to integrate multi-domain datasets and facilitate the coupling simulations across a variety of disciplines. The target users of the proposed automated workflow include urban researchers, operational scientists, and software developers with domain expertise and either entry-level or intermediate computer science and programming skills. Detailed design requirements of our automated workflow are as follows: 

\begin{itemize}
  \item Compatibility and Portability: The workflow should be designed as a generalized workflow, implementable in popular programming languages like Python. It should be modular and flexible, enabling deployment and execution on widely-used online platforms such as Jupyter Notebook. 
  \item Automated Knowledge Source Acquisition: The workflow should be integrated with online databases and archives for scientific research literature via APIs. This integration will enable the agent to automatically download relevant research papers and technical manuals based on user-defined keywords related to a specific scope, dataset, or simulation model.
  \item Autonomous Knowledge Extraction: The workflow should be capable of comprehending human natural language, recognizing domain-specific terminologies, such as identifying concepts as entities and categories as classes, and understanding the relationships between various concepts and classes.   
  \item Automated Ontology Design: The workflow should adhere to the practices and procedures of well-established ontology design methodologies, including Methontology, OTKM, and NeON (see Section \ref{Ontology Design and Development}). This adherence will enable it to create robust knowledge representations characterized by high granularity, reusability, flexibility, and semantic consistency.
  \item Ontology Interoperability: The workflow should be able to export ontologies as knowledge graphs or in widely accepted ontology languages, such as OWL and RDF (see Section \ref{Semantic Web Standards}). This functionality will enable the ontologies to be interpreted by popular ontology management software tools, facilitating validation processes and the creation of database schema for data modeling and integration. 
  \item Self-validating: The autonomous \rev{tool} should have the capability to validate the ontologies it generates by referencing existing domain ontologies or relevant literature. This process should include checking the definitions of concepts, relationships, and semantic mappings to ensure accuracy and consistency.
\end{itemize}

\rev{Our proposed method}, designed and developed based on the aforementioned requirements, will later be employed to create a sample \rev{scientific} ontology as a proof-of-concept. This will involve integrating emerging datasets and simulation outputs from the intermodal transportation sector, including the Freight Analysis Framework (FAF) and Freight Transportation Optimization Tool (FTOT). This integration aims to facilitate informed and comprehensive decisions to optimize complex intermodal freight transportation systems. More details regarding the data and simulation used in this work are provided in Section \ref{sec:datamodels}.


\subsection{Methodological Foundation} \label{sec:sonstructing_pntology}
\begin{figure}[!htpb]
    \centering
    \includegraphics[width=\textwidth]{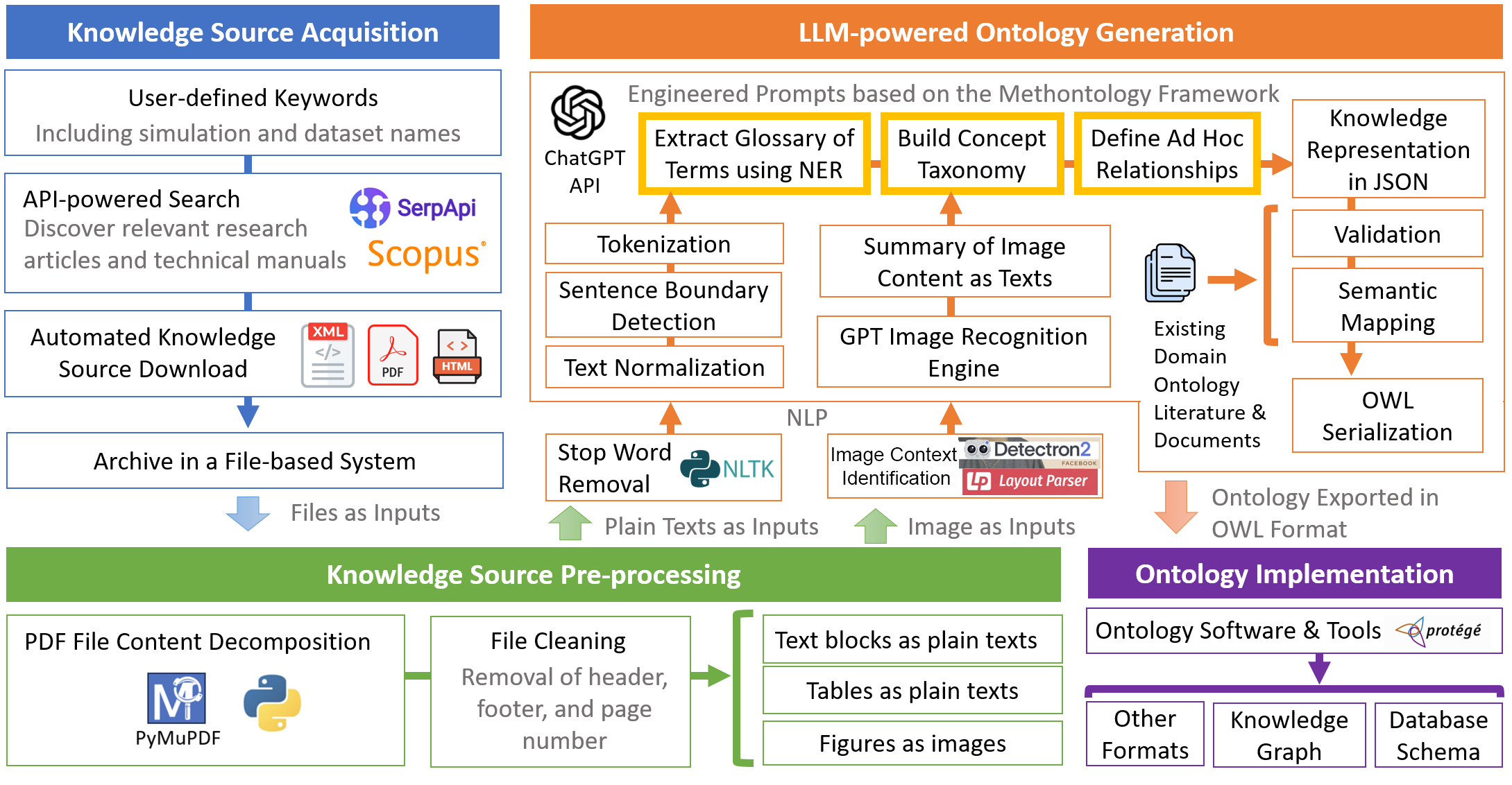}
    \caption{ \rev{Overall architecture of the applied workflow for ontology generation with four major modules }}\label{fig:Infographic2}    
\end{figure}

The conceptual design of our autonomous workflow is depicted in Figure \ref{fig:Infographic2}, which comprises four primary modules: (a) Knowledge Source Acquisition, (b) Knowledge Source Pre-processing, (c) LLM-Powered Ontology Generation, and (d) Ontology Implementation. Each component encompasses a series of procedures implemented using Python libraries. Additionally, the entire process is \rev{based in Python, combining JupyterLab and the use of scripting for automated orchestrated execution of the process. This facilitates human supervision of the workflow step-wise.} Detailed workflow and procedures within individual modules are presented through the following subsections.

\subsubsection{Knowledge Source Acquisition and Pre-Processing}
The knowledge source acquisition module automates the discovery, downloading, and archiving of digitized files (e.g., PDFs and XMLs). These files contain essential domain knowledge and are retrieved based on user-defined keywords. These keywords may refer to specific urban or environmental datasets, scientific simulations, and more. It employs the Scopus API, an academic database that discovers and downloads research articles using these keywords as search criteria. This API-powered component searches the internet for resources that match the search criteria and downloads them. The resources, including research articles and technical manuals for datasets or simulation models, are saved in formats like PDF, HTML, and XML. They are then stored in a designated file directorate, which other components of our automated workflow can access.
 
The knowledge source pre-processing module is designed to prepare digital files, previously downloaded, into formats suitable for input into the LLM for ontology generation. This module integrates content analysis and data extraction libraries with custom Python scripts to initially cleanse the digital files. This cleansing process removes unnecessary components such as headers, footers, page numbers, and special characters, which could potentially impair the performance of subsequent natural language processing tasks and the LLM's. Subsequently, the content is extracted and transformed into plain text (for text blocks and tables) or images (for figures), depending on their original format. The extracted content is then divided into subsets that comply with the word count limitations of the LLM for each request, specifically using the ChatGPT API in this study.

\subsubsection{LLM-powered Ontology \rev{Construction} - Input Preparation}
The LLM-powered ontology generation module leverages ChatGPT's natural language processing (NLP) and reasoning capabilities to automatically generate scenario-based ontologies. This process is facilitated through the use of prompt templates, which are developed based on the core principles of the Methontology framework. 

Before initiating the ontology development procedures, plain text and image inputs are processed through two separate workflows to enhance input data quality. Text inputs undergo a series of NLP procedures, including tokenization, stop word removal, sentence boundary detection, and text normalization. Concurrently, images are processed through the GPT image recognition engine via API to extract embedded text and summarize the content in text form. The refined text inputs from both plain texts and images are then fed into the ChatGPT-4 API, which serves as the central reasoning engine, to construct the ontology. The Methontology framework's procedures have been translated into structured prompts that utilize the flexibility of Python scripting and the ChatGPT API, creating a maintainable workflow. This workflow instructs the LLM on how to create scientific ontologies based on pre-processed input data. Detailed data flow and its procedure for processing the plain text inputs are described as the following:
 \begin{enumerate}  
    \item Stop Word Removal: a technique used in many traditional NLP tasks where common words such as ``and", ``the", ``is", etc., which are deemed to have little meaning, are removed from the text. This is typically done to reduce the dimensionality of the data and focus on more meaningful words for tasks like text classification, keyword extraction, and more, and is conducted outside the ChatGPT API using the NLTK library. 
    \item An NLP procedure that cleans and standardizes text by converting all characters to lowercase, removing or replacing special characters and punctuation, standardizing dates, numbers, and other non-standard text forms, and managing whitespaces. This process ensures uniformity in the text, minimizing variability that could detract from analysis. Text normalization is integrated into the LLM's functionality via the ChatGPT API.
    \item Sentence Boundary Detection: Following text normalization, this process identifies the start and end points of sentences within the text. Accurate sentence boundary detection is essential for understanding the text's structure and for facilitating subsequent per-sentence processing. It segments a large text into manageable, analyzable units. This functionality is also incorporated into the LLM via the ChatGPT API.
    \item Tokenization: Following sentence boundary detection, the text within each sentence is tokenized. This process splits the text into tokens—typically words, phrases, or other meaningful elements. Tokenization transforms the text into discrete units suitable for further processing, such as parsing, part-of-speech tagging, or input to LLMs for reasoning, interpretation, and knowledge extraction. This functionality is also integrated into the LLM through the ChatGPT API.
\end{enumerate}

Detailed data flow and its procedure for processing the image inputs are described as the following:
\begin{enumerate}  
     \item \rev{Image Context Recognition: The process uses Optical Character Recognition (OCR) technology to detect and convert text surrounding images into machine-readable characters. This includes extracting annotations, axis labels, data points, and descriptive text essential for interpreting the graphical content. This process is conducted outside the ChatGPT API using Layout Parser \cite{shen2021layoutparser} and the Detectron2 model \cite{wu2019detectron2} for image separation serves as component identification to filter those that represent figures for further processing.}
     
     \item GPT Image Recognition Engine: a component of a larger AI system that utilizes models trained to recognize and interpret content within images. This engine processes images to identify objects, features, and patterns, often using advanced neural networks optimized for image analysis. In the context of research articles, this involves recognizing graphical data, text within figures, or even the structure of the figures themselves. This functionality is integrated into the LLM through the ChatGPT API.
     
     \item Summary of Images as Texts: The process synthesizes key information from an image into a concise textual summary. It uses AI to interpret recognized objects, distilling these elements to highlight the essential insights or data presented. This is especially useful in academic settings for quickly conveying the significance and results depicted in complex figures or diagrams. This functionality is also integrated into the LLM through the ChatGPT API.
\end{enumerate}

\subsubsection{LLM-powered Ontology \rev{Construction} - Knowledge Extraction}
This section details the workflow and procedures for leveraging the ChatGPT API's language understanding and reasoning capabilities, as highlighted by the light-orange border in Figure \ref{fig:Infographic2}, to develop the ontology. The ontology generation process relies on a series of engineered promotes that are derived and guided by the ontology design procedures from the Methontology framework to guide the LLM in performing specialized tasks. Currently, we do not apply any prompt tuning or fine-tuning technique to our method.
\begin{enumerate}  
     \item Extracting a Glossary of Terms using Named Entity Recognition (NER): This initial step in ontology creation employs NER, a pivotal NLP technique for identifying and categorizing key textual information. The process involves scanning text to detect and label critical domain-specific concepts as entities. These entities form the foundational elements of the ontology, enhancing semantic analysis by clearly outlining the terms and their relationships within the domain.
     \item Build Concept Taxonomy: A step focuses on organizing and structuring the identified concepts into a hierarchical framework. Building a concept taxonomy involves arranging concepts (or classes) identified during the conceptualization phase into a taxonomic (hierarchical) structure based on ``is-a" relationships. This hierarchy represents a superclass-subclass relationship between broader and more specific concepts.
     \item Define Ad Hoc Relationship: This step is a crucial phase in the ontology development process. It involves identifying and specifying the relationships that exist between the concepts that are specific to the domain being modeled and do not necessarily fit into well-defined hierarchical (is-a), associative (part-of) relationships, \rev{ or attribute (has property) relationships.} 
\end{enumerate}

After creation, the scenario-based ontologies generated by our proposed methods undergo validation against the documentation of existing domain ontologies, which typically describe high-level, generic domain knowledge. This validation is conducted using the reasoning capabilities of LLMs. This process aims to ensure accurate and relevant identification of domain concepts, as well as logical mapping of relationships between concepts and classes. Additionally, this procedure ensures semantic consistency across the AI-created ontology through a semantic mapping process that leverages the existing domain ontology. After these procedures, the finalized ontology output is then stored in an OWL format, which is interoperable and compatible with popular ontology management software tools and modern semantic web technologies.

The OWL ontology produced by our methods is subsequently imported into widely-used ontology management software, \textit{Protégé} as part of an existing project. This process supports a range of use cases that enhance the development of decision support systems or digital twins by providing direct or indirect pathways to generate essential software components. Examples of these components include data models and queries, knowledge graphs, and ontologies stored in other popular Semantic Web formats \citep{dong2007application}.

A specific use case involves converting the LLM-generated OWL ontology into SPARQL using Protégé. SPARQL is primarily used to query and manipulate data stored in the RDF format \citep{kumar2013querying}. This capability enables decision support applications to more effectively discover, retrieve, and integrate data resources from RDF databases. In this setting, the ontology can be used to facilitate data management from relational database \citep{lubbad2018ontology}, enabling a data provision pipeline to support data-driven insights during the development of urban decision support systems. 


\subsubsection{\rev{Ontology Evaluation}}
\rev{It is crucial to highlight the current empirical evidence that underscores the robust capabilities of applying methodologies akin to those outlined in \cite{fernandez1997methontology}, now seamlessly executed through Large Language Models (LLMs). The human-crafted ontology amalgamates inputs from undisclosed sources, revealing the logical interconnections among the presented classes and subclasses. Conversely, the data utilized for the automated pizza ontology stems from a distinct singular source, presented in a text format through the PDF files. The modular components in pizza source document provide a systematic breakdown of information, which is then processed through the proposed framework and LLM. The automated pizza ontology, showed decomposed in an outer \footnote{ Available at \url{https://github.com/jtupayachi/Auto_Ontology/blob/main/figures/SECTION1_PIZZA.png}} and inner \footnote{Available at \url{https://github.com/jtupayachi/Auto_Ontology/blob/main/figures/ONTOLOGY_TOPPINGS.png}} sections mirror the logical components and properties resulting from the method.}

\rev{Ontologies, define a subclass-of relationship. This relationship, could be expressed as ``A is-a B," signifies A represents a more specific or specialized category than B. In essence, this arrangement delineates the hierarchical structuring of knowledge. We evaluate the feasibility of our methodology through a comparative analysis, juxtaposing our AI-generated ontology with the widely recognized Pizza Ontology (Human-crafted). The work discussed by \cite{rector2004owl} depicts the challenges faced in the domain of web ontology language description and generation logic. These challenges often can lead to conceptual misunderstandings and underscore the necessity for expert domain knowledge to effectively structure, comprehend, and ultimately construct knowledge representations. The use of an automated ontology creation process requires qualitatively verifying its proper generation, which contrasts with the Pizza representation, to ensure the LLM-generated ontology is valid. The human-created ontology \footnote{Available at: \url{https://github.com/jtupayachi/Auto_Ontology/blob/main/comparison_pizza/pizza.owl}}, shows an ``is-a" and ``subclass-of" relationships. This subtype or subclass indicates a specialization-generalization relationship. An automated ontology \footnote{Available at: \url{https://github.com/jtupayachi/Auto_Ontology/blob/main/comparison_pizza/automated.rdf}} per section is fully generated via the LLM. This ontology depicts vegetable toppings embedded as individuals in the LLM-generated model. The inclusion of individual instances composed of artichokes, mushrooms, onions, and tomatoes resembles the human-made ontology. The automated ontology modules generated (one per each text section - topic) were merged using the OWL API.}

\rev{Similarly, we represented the inner components of human-crafted ontology \footnote{Available at: \url{https://github.com/jtupayachi/Auto_Ontology/blob/main/figures/onto_outer.png}}, through the main pizza class and its corresponding subclasses, which encompass styles, characteristics, and components. Additionally, a deeper level of the labeling under 'Pizza' \footnote{ Available at: \url{https://github.com/jtupayachi/Auto_Ontology/blob/main/figures/onto_inner.png}}
 provides a broader context and a more general understanding of this topic.} \rev{Several methods have been proposed to evaluate ontology outcomes and assess the alignment of frameworks and methods with the intended topic. \cite{redmond2011computing} argue that direct textual comparison becomes ineffective in one-to-one comparisons. They focus instead on verifying and mapping axioms and target ontologies using a differential engine built on the OWL API, which calculates structural differences between ontologies. Similar approaches are followed by other researchers, such as \cite{malone2010modeling} with the Bubastis tool and OWLDiff by \cite{kremen2011owldiff}. It is important to note that these methods are particularly suited for scenarios where ontologies undergo modifications, aiming to uncover such differences effectively.}

\rev{A more suitable approach for comparing our framework with a human-crafted ontology involves using CQs \cite{kommineni2024human}. The approach utilizes pre-defined CQs, such as "Write the competency questions based on the abstract level concepts for describing ...", to evaluate the utility of an ontology. Qualitative comparisons between ontologies created using different methods can also be conducted by examining the ontology's capability to address individual CQs. \cite{noy2002evaluating} proposed a CQs-based method that focuses on merging and aligning ontologies, where the authors highlight alignment points such as "Industrial org" and "Commercial Organization". Although these terms differ, they represent similar concepts. This method relies on user inputs, which, when combined with the PROMPT tool \citep{noy2003prompt}, can be used to compare and evaluate the structure within each ontology. The evaluation considers the differences in the Prompt-Diff table, specifically examining the number of rows added, deleted, or changed in their operation or mapping-level columns. Other studies have also introduced various approaches for evaluating the creation of ontologies using CQs, similarly as used in \cite{kommineni2024human}}. 
\begin{table}[h!] \caption{Qualitative pizza ontology comparison using CQs and qualitative analysis. A SPARQL platform is employed for the execution of the queries and interface with the (.rdf or .owl) files. The keyword ``ontology'' in this table references the url of the uploaded ontological files.}
\centering
\small
\renewcommand{\arraystretch}{1.2}
\setlength{\tabcolsep}{4pt}
\begin{tabular}{|>{\centering\arraybackslash}m{1.5cm}|>{\centering\arraybackslash}m{4cm}|>{\centering\arraybackslash}m{3cm}|>{\centering\arraybackslash}m{4cm}|>{\centering\arraybackslash}m{3cm}|}
\hline

\textbf{CQ} & \multicolumn{2}{c|}{\textbf{Queries on Human-Crafted Ontology }} & \multicolumn{2}{c|}{\textbf{Queries on Automated Ontology  }} \\
\hline
What are the primary categories within the pizza ontology? & 
\textbf{PREFIX:} rdf: \url{<http://www.w3.org/1999/02/22-rdf-syntax-ns#>}

    SELECT DISTINCT ?class
WHERE \{
  ?class rdf:type owl:Class .
\}
& 
Pizza, 

PizzaBase, 

Food, 

Spiciness 

\color{blue}{105 in total}
& 
\textbf{PREFIX} rdf: \url{<http://www.w3.org/1999/02/22-rdf-syntax-ns#>}
 SELECT DISTINCT ?class WHERE \{ ?class rdf:type ontology:Class .\}
& 
Food, 

Process, 

Business, 

Culture 

\color{blue}{34 in total}
\\
\hline
What are the main ingredients of a pizza? &
Not found in ontology
& 
Not found in ontology
& 
\textbf{PREFIX} rdf: \url{<http://www.w3.org/1999/02/22-rdf-syntax-ns#>}, \textbf{PREFIX} rdfs: \url{<http://www.w3.org/2000/01/rdf-schema#>}
 SELECT ?individual ?label WHERE \{ ?individual rdf:type ontology:Ingredients . OPTIONAL \{ ?individual rdfs:label ?label \}\}
& 
Cheese, 

Mozzarella, 

Tomatoes 

\color{blue}{18 in total}
\\
\hline
What are the different variations of toppings used for a pizza? &  
\textbf{PREFIX} rdf: \url{<http://www.w3.org/1999/02/22-rdf-syntax-ns#>}, \textbf{PREFIX} rdfs: \url{<http://www.w3.org/2000/01/rdf-schema#>}
 SELECT ?class

WHERE \{

  ?class rdfs:subClassOf ontology:VegetableTopping .

\}

& 
TomatoTopping, 

HotGreenPepperTopping, 

JalapenoPepperTopping,

ArtichokeTopping,

AsparagusTopping,

OnionTopping,

PeperonataTopping,

CaperTopping,

OliveTopping 

\color{blue}{22 in total}

& 
\textbf{PREFIX} rdf: \url{<http://www.w3.org/1999/02/22-rdf-syntax-ns#>}, 

\textbf{PREFIX} rdfs: \url{<http://www.w3.org/2000/01/rdf-schema#>}
SELECT ?subclass WHERE \{ ?subclass rdfs:subClassOf ontology:PizzaToppings .\}
& 
CheeseToppings, FruitToppings, HerbAndSpiceToppings, MeatToppings, OtherToppings, SauceToppings, SeafoodToppings, VegetableToppings
\\
\hline
Which dishes use the dough as a main ingredient? &  
Not found in ontology
& 
Not found in ontology
& 
\textbf{PREFIX} rdf: \url{<http://www.w3.org/1999/02/22-rdf-syntax-ns#>}
 SELECT DISTINCT ?dish WHERE \{ ?dish rdf:type ontology:Dish . ?dough rdf:type ontology:Dough .\}
& 
Calzone, 

Pizza
\\
\hline
What are the Vegetable Toppings? &  
\textbf{PREFIX} rdf: \url{<http://www.w3.org/1999/02/22-rdf-syntax-ns#>}, 


 SELECT ?individual
WHERE \{
  ?individual 
              rdf:type ontology:VegetableToppings .
\}




& 
Tomato,

Hot Green Pepper,

Jalapeno Pepper,

Artichoke,

Asparagus,

Onion 

\color{blue}{22 in total}

& 
\textbf{PREFIX} rdf: \url{<http://www.w3.org/1999/02/22-rdf-syntax-ns#>}
 SELECT ?vegetableTopping WHERE \{ ?vegetableTopping rdf:type ontology:VegetableToppings .\}
& 
Artichokes, 

Mushrooms, 

Onion, 

Tomatoes
\\
\hline
What are the different types of Pizzas? &  
\textbf{PREFIX} rdf: \url{<http://www.w3.org/1999/02/22-rdf-syntax-ns#>}, 

\textbf{PREFIX} rdfs: \url{<http://www.w3.org/2000/01/rdf-schema#>}

   SELECT ?pizzaType ?label

WHERE \{
  ?pizzaType rdf:type owl:Class .
  ?pizzaType rdfs:subClassOf* ontology:Pizza .
  OPTIONAL \{ 
    ?pizzaType rdfs:label ?label .
  \}
& 
AmericanHot, 

Cajun,

Capricciosa,

Caprina,

Fiorentina,

FourSeasons

\color{blue}{25 in total}
& 
Not found in ontology
& 
Not found in ontology
\\
\hline
\end{tabular}
\label{tab:pizza_ontology}
\end{table}

\subsubsection{\rev{CQ-based Qualitative Analysis}}
\rev{Unlike many existing studies that focus on generating foundational domain ontologies and knowledge representations \citep{kommineni2024human}, our method expands existing domain ontologies to create scenario-based ontologies from new datasets or simulations, facilitating practical decision support solutions. To evaluate the AI-generated ontology from our method, we adopt the commonly accepted ontology evaluation methods using CQs and qualitative analysis based on justifications from human-expert \citep{bezerra2013evaluating, ren2014towards}. Table \ref{tab:pizza_ontology} presents a side-by-side comparison of the AI- and human-generated pizza ontologies, and it includes the formulation of CQs and the SPARQL queries used for evaluation. The results indicate that the outputs of the AI-generated ontology are sensible as with those from the human-crafted ontology. }

\rev{In our scenario, it is important to note that the primary data source for creating the pizza ontology is not identical to the one used for the human-crafted ontology. This discrepancy may lead to variations in scope, classes, and subclasses, as well as the resulting SPARQL queries. To perform a refined evaluation, it is crucial for the initial sources of both approaches to be identical. As shown in Table \ref{tab:pizza_ontology}, the content from the AI-generated ontology is queried using SPARQL and provides logical and reasonable responses to address each CQ. Although there may be some differences in the actual content, the outputs demonstrate that the classification and alignment of the AI-generated pizza ontology are sensible, logical, and practical. Our comparative analysis ensures that the AI-generated ontology can accurately reflect knowledge without providing hallucinated or fabricated results.
}


\section{Case Study: Optimizing the Intermodal Freight Transportation System} 
\label{sec:Future_Discussion}
In this section, we present a compelling case study to demonstrate the application of our methodology in addressing a real-world decision optimization challenge. Our focus is on the analysis of intermodal freight transportation across the United States, where we leverage the integration of multiple datasets. The primary objective of the study is to optimize this complex system with the goals of enhancing fuel efficiency and, at the same time, reducing Greenhouse Gas (GHG) emissions and total operation costs. This endeavor stands as a demonstration of the capability of our methodology in addressing multifaceted, real-world challenges with tangible environmental and economic implications in urban systems.

\begin{figure}[!htp]
    \centering
    \includegraphics[width=\textwidth]{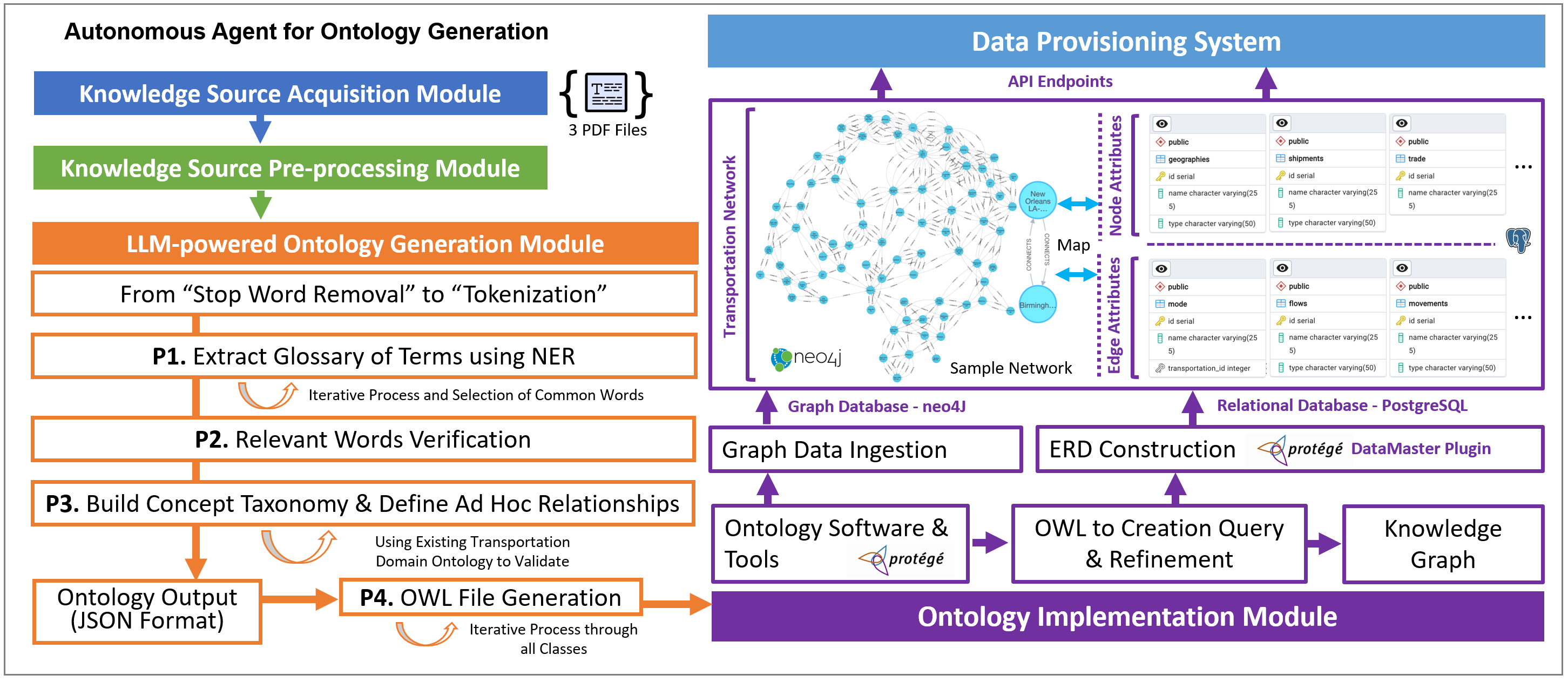}
    \caption{ \rev{The demonstration includes prompt templates used in the LLM-powered workflow for constructing ontology and knowledge, as well as the design of the data provisioning system.     
    }}\label{fig:LLM_process}    
\end{figure}

\subsection{Data Sources and Simulation Models} 
\label{sec:datamodels}
Our methodology integrates three primary data sources (FAF and FTOT) alongside an academic source this enriches the knowledge and analysis of intermodal transportation networks as represented under Figure \ref{fig:Infographic2}. The Freight Analysis Framework (FAF) \citep{faf} and its accompanying manual serve as foundational resources, identifying crucial freight transportation hubs across diverse regions. Identified hubs provide a framework for mapping significant geographic and logistic parameters that define major nodes within transportation networks. Similarly, the FTOT \citep{ftot} and documentation are instrumental in permitting the creation of the intermodal network. Calculations under the criteria of the shortest path; this transportation network is represented via shapefiles. The generated ontology will allow a decision support program, script, or LLM, to perform calculations and handle services such as an endpoint REST API to compute unique routes for various nodes, encompassing road, rail, and water modes, while delineating the properties and relationships within these networks. Manuals ensure a comprehensive understanding of the methodologies and algorithms applied in these tools, enhancing the accuracy of simulations and optimizations in our model. Academic contributions, such as methodological approaches for calculating equivalent CO$_2$ emissions, among other sources, enhance our understanding and integration of emissions considerations within the LLM framework, further refining our strategies for optimization and sustainability in intermodal freight transportation. In addition to the aforementioned primary data and simulations encompassing operational costs, fuel consumption, and GHG emissions across various modes of freight transportation in the US, our approach integrates the comprehensive array of intermodal GIS transportation networks provided by FTOT. The location of major critical transportation hubs is documented through FAF.


\rev{\subsection{Ontology and Knowledge Graph Construction}\label{sec:OKGC} }

\rev{We applied our LLM-based methods to the three documents of the Freight Analysis Framework (FAF) and the Freight and Transportation Optimization Tool (FTOT), deriving an OWL ontology. This ontology was then converted into a knowledge graph in RDF format to support more practical data integration and management applications.}

\begin{figure}[!htp]
    \centering
    \includegraphics[width=\textwidth]{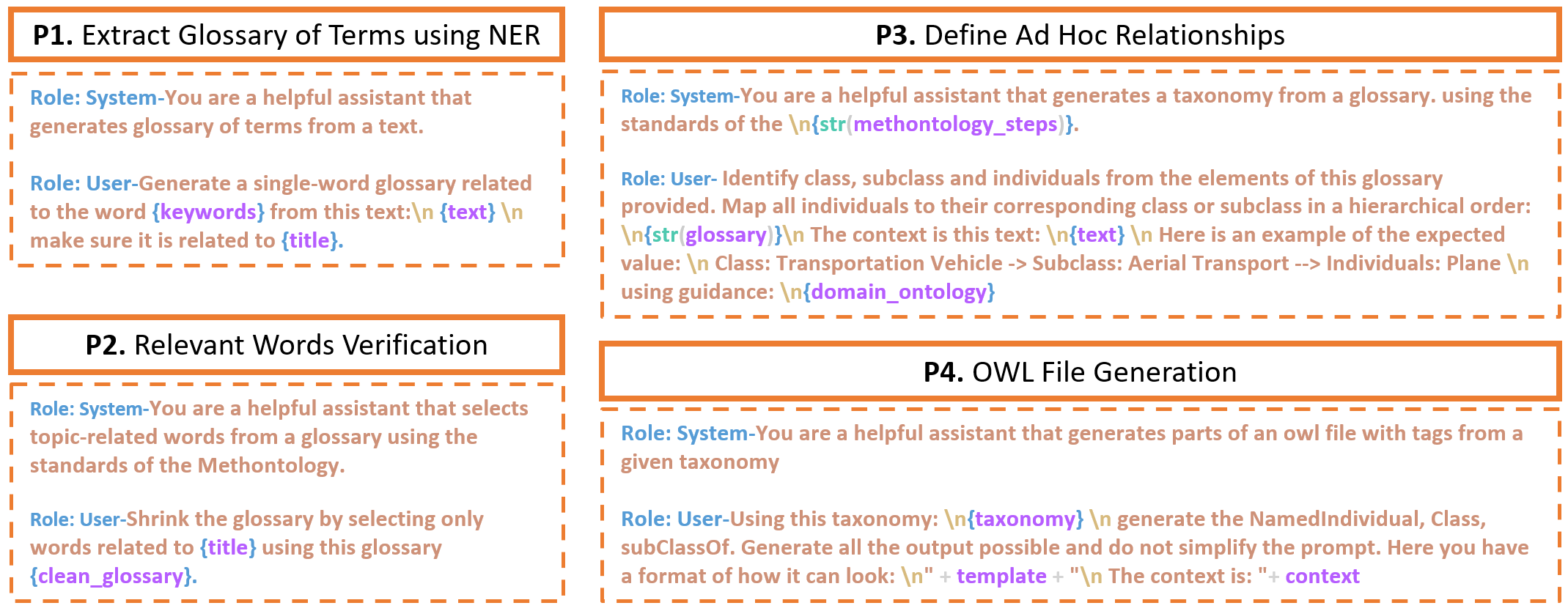}
    \caption{ \revv{Demonstration of prompt templates used to instruct the LLM.     
    }}\label{fig:LLM_process_pt}    
\end{figure}

\rev{Figure \ref{fig:LLM_process} illustrates the implementation details, including knowledge representations and database components. \revv{The instructions and prompt templates provided to the LLM for generating the ontology at each critical step are demonstrated in Figure \ref{fig:LLM_process_pt}}. From the supplied PDF files, we perform the cleaning phase, which includes stop word removal, punctuation, and special character handling.
Next, the process relies on glossary generation, a crucial step where the text is simplified into a set of relevant words. This step is supported by an iterative validation process that enhances robustness and filters out words that do not persist through multiple iterations. We supply the elaborated prompt with the \texttt{title} and \texttt{keywords} of the provided section and request the LLM for confirmation on the content it provides. After the completion of this iterative process, the LLM moves to the next function, which verifies consistency and ensures that the relevant words glossary are aligned with the section. This glossary serves as the input for taxonomy generation, where we establish the relationships and hierarchical order within the \texttt{glossary} of words and concepts. Our OWL file generator then transforms the identified classes, subclasses, and individuals into a readable OWL file. Following this, an iterative process ensures the LLM finalizes the output until the stop condition in the metadata is met. While challenges may arise, different methods can be applied to address them. The LLM is then used to transform the OWL output and construct a query for the creation of a table schema, later represented as the entity-relationship diagram (ERD). More technical details on how the ERD is implemented using the AI-generated ontology are provided in Section \ref{sec:onton-data}.}
The AI-generated intermodal ontologies is composed of three separate ontology modules:
\rev{
\begin{enumerate}
\item The first module is the ontology supporting the FAF data, providing essential data descriptions for developing a data provisioning system. This system encompasses factors such as types of transportation, cargo deliveries, geographical areas, and data sources for transportation data.
\item The second module includes the FTOT network, which incorporates the spatial data (GIS data) necessary for calculating routes and inter-hub distances.
\item The third module comprises the problem statement and optimization models derived from various academic resources. This module focuses on greenhouse gas emission calculations, optimization formulations, and intermodal transportation agents.
\end{enumerate}
}

\begin{figure}[!htp]
    \centering
    \includegraphics[width=\textwidth]{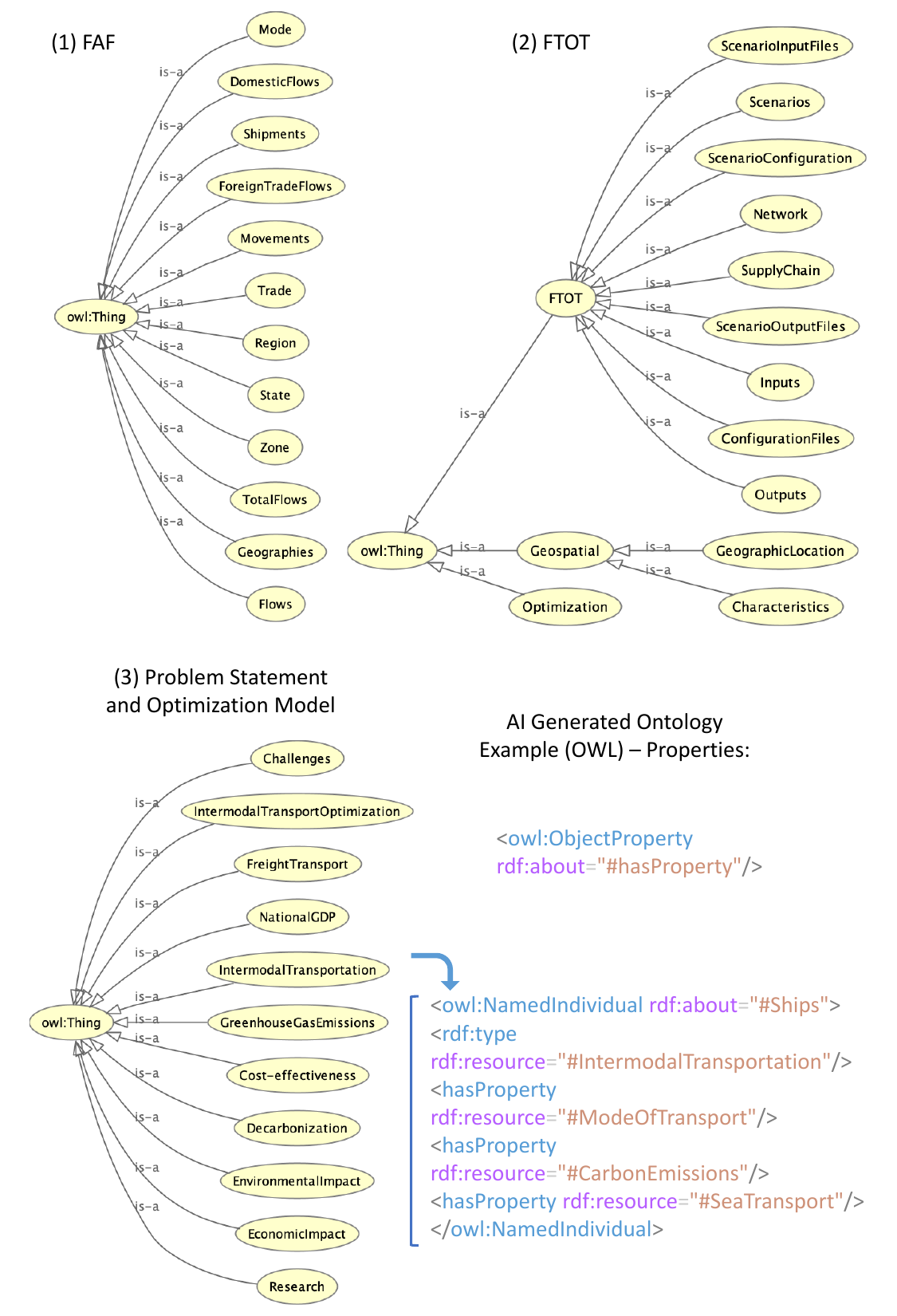}
    \caption{  \rev{AI-generated ontology for describing concepts, data, and simulation outputs for optimizing intermodal freight transportation.} }\label{fig:freight_transportation_ontology}    
\end{figure}

\rev{
To formulate the decision support workflow and problem-solving strategies, we use the third ontology module to derive two attributes of the core concepts involved in optimizing intermodal freight transportation. These attributes include 'properties,' which indicate the connections and relationships between different concepts, and 'individuals,' which are specific instances created for particular objects or entities to define core decision support metrics. Demonstration of the AI-generated ontology and sample attributes are provided in Figure \ref{fig:freight_transportation_ontology}.
}

\rev{\subsection{Data Provisioning System: Technologies and Implementation}\label{sec:DPSDI}}
\rev{A robust data provisioning system plays a critical role in the discovery, querying, and retrieval of data necessary for supporting informed decisions. In our case study, we utilize an AI-generated ontology and knowledge graph to guide the design and development of our data provisioning system. This system consists of essential databases for storing intermodal transportation data and simulation outputs, as well as web API endpoints for querying and accessing these datasets. Given the nature of the FAF and FTOT data, which encompass large transportation networks (topologically connected graph data) connecting numerous origins and destinations (e.g., intermodal transits and warehouses in major US cities) using different transportation modes and routes, we employ both graph database and a relational database to store the integrated intermodal transportation attributes. }  

\rev{For the graph database management system, we used neo4j to store the FAF transportation network as graph data elements (nodes, edges connecting them, and attributes of nodes and edges). To import the FAF network into a neo4j database, we derived the transportation network based on the Origin-Destination (OD) pairs documented in the \revv{FAF data.} The resulting network (represented by a graph) describes the simplest form of the connectivity of roads, rails, and rivers across the US, with graph components (e.g., nodes as cities and edges as routes) serving as building blocks to join attributes and metrics necessary for decision optimization. The primary purpose of the neo4j database is to identify all possible combinations of modes and routes (represented by lists of nodes and edges) based on any user-defined OD pairs requested through web APIs.
To optimize neo4j's performance in traversing the network, we store only the transportation network and its topology within the neo4j database, including only essential attributes (e.g., node and edge IDs, modes, and edge distances and slopes). These essential attributes are added to the graph to allow the neo4j database to apply constraints during the network traversal. Attributes related to the route (edge) geometry are pre-calculated using spatial analysis tools and included in the graph as numerical attributes. Additional graph element attributes pertaining to nodes and edges, which characterize different facets of intermodal transportation as decision metrics, are not imported into the graph database. Instead, these non-essential attributes are stored in the relational database. This practice of partitioning non-essential attributes from the graph aims to enhance neo4j’s performance in navigating the large-scale transportation network.}

\rev{Once all alternative combinations of modes and routes are identified from the neo4j database, we use the lists of node and route IDs for each combination to retrieve the necessary data and simulation outputs as decision metrics from the relational database. In this study, we use PostgreSQL, with its schema guided by the knowledge graph, to integrate and store vast and diverse urban data and simulation outputs (e.g., FTOT and URIs to traffic data APIs) that characterize intermodal transportation processes. Examples of these metrics include total GHG emissions, total shipment time, total fuel consumption, and total operation cost. These metrics are either pre-calculated and stored in the relational database or dynamically calculated using real-time traffic information (e.g., congestion indexes and road closure information) retrieved from third-party data APIs through time-scheduled scripts deployed in a cloud-based server environment.
The AI-generated ontology guides the design of the relational database schema and the information searches in the relational databases based on user-defined criteria and requirements (e.g., inclusion or exclusion of GHG emissions) for optimization.
}


\rev{
\subsection{Data Provisioning System: Database Schema Design}
\label{sec:onton-data}
}
\rev{
The ontology and its knowledge graph are primarily used to construct the database schema and Entity Relationship Diagram (ERD) for the PostgreSQL database, as well as to provide essential metadata and descriptions for each metric used in decision-making. By traversing the graph stored in neo4j, we can identify all possible routes between a set of user-defined origins and destinations. Based on these identified routes, our ontology facilitates the discovery and retrieval of all required metrics from the PostgreSQL database by mapping individual metrics to their associated graph components in the identified routes, such as edges and nodes. In terms of implementation, our proposed LLM-powered methods can export the ontology of intermodal transportation data and simulations, such as FAF and FTOT, in OWL format. Many existing ontology management software tools provide straightforward methods and plugins for converting OWL to relational database schemas \citep{mogotlane2014semantic}. In our case study, we use Protégé and its DataMaster plugin to convert the AI-generated ontology into the PostgreSQL database schema. DataMaster allows users to import database schemas into an ontology and vice versa \citep{mogotlane2016automatic}. If needed, custom scripts in Python or Java can also be created to translate the ontology and knowledge graph into a relational schema. Libraries such as RDFLib (for Python) or Apache Jena (for Java) can assist with parsing and processing OWL files.}


\begin{figure}
    \centering
    \includegraphics[width=1\textwidth]{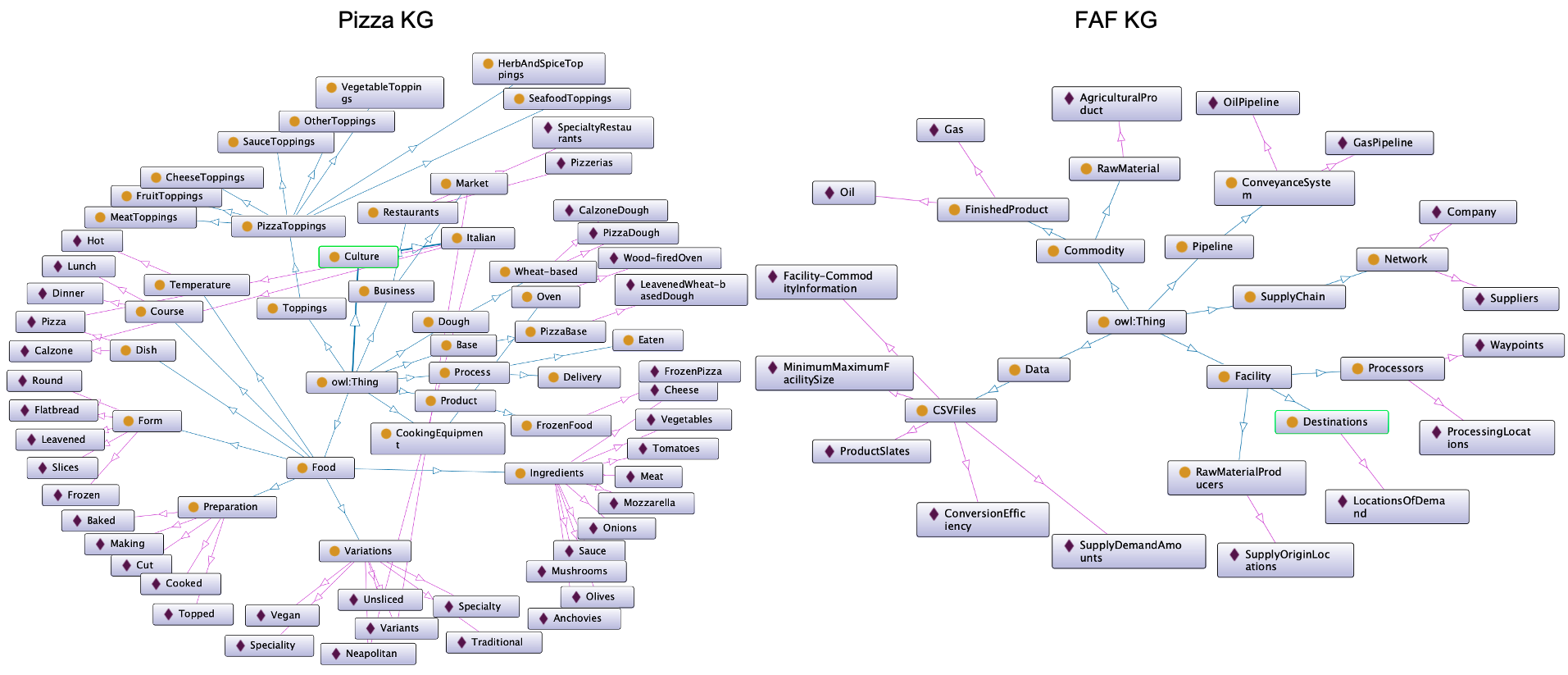} 
    \caption{\rev{Automated Knowledge Graph Representation}}
    \label{fig:KG}
\end{figure}

\rev{A subset of the ERD for the relational database is depicted in the purple boxes in Figure \ref{fig:LLM_process}. The sample subset of the FAF network data stored in the neo4j database is shown in Figure \ref{fig:LLM_process}. Both databases provide web API endpoints as URIs to facilitate information discovery, query, and retrieval through common internet communication protocols, such as REST. The client-side application of a decision support system can send requests to these URIs, which are later manually added to the knowledge graph to enhance data observability. Under Figure \ref{fig:KG}, we present a snippet of the knowledge graph that includes both the generated pizza knowledge graph and the FAF dataset knowledge graph. This represents an intermediate step between the conceptual ontology and the creation of relational databases. We utilized the OntoGraf plugin from Protégé for visualization.}\\

\begin{table}[h!]\caption{\revv{Ontology-guided selection of decision support metrics and workflow using FTOT and FAF ontology. }}
\label{tab:FAF_ontologyUsecause}
\centering
\small
\renewcommand{\arraystretch}{1.2}
\setlength{\tabcolsep}{4pt}
\begin{tabular}{|>{\centering\arraybackslash}m{1.5cm}|>{\centering\arraybackslash}m{5cm}|>{\centering\arraybackslash}m{4cm}|>{\centering\arraybackslash}m{3cm}|>{\centering\arraybackslash}m{1cm}|}
\hline
\textbf{CQ} & \multicolumn{2}{c|}{\textbf{Automated Ontology Prefix, Query and Output }} & \textbf{Query Output Description} &{\textbf{Dataset}} \\
\hline
What are the primary scenario parameters within the FTOT Ontology? & 
\textbf{PREFIX:} rdf: \url{<http://www.w3.org/1999/02/22-rdf-syntax-ns#>}

\textbf{PREFIX:} rdfs: \url{<http://www.w3.org/2000/01/rdf-schema#>}

\textbf{PREFIX:} owl: \url{<http://www.w3.org/2002/07/owl#>}
    SELECT ?parameter
WHERE \{
    ?parameter rdf:type base:ScenarioParameters .
\}



& 
Scenario\_Name, Scenario\_Description, RMP\_Commodity\_Data, Destinations\_Commodity\_Data, Disruption\_Data,



\color{blue}{20 in total}
& 
The parameters, metadata related parameters, details on raw material producers, destinations, and information on disruption events.
& 
FTOT
\\

\hline
What are the primary scenario inputs within the FTOT Ontology? &

\textbf{PREFIX:} rdf: \url{<http://www.w3.org/1999/02/22-rdf-syntax-ns#>}

\textbf{PREFIX:} rdfs: \url{<http://www.w3.org/2000/01/rdf-schema#>}

\textbf{PREFIX:} owl: \url{<http://www.w3.org/2002/07/owl#>}

SELECT ?input
WHERE \{
    ?input rdf:type base:ScenarioInputs .
\}



& 
geospatialinformation,

networkattributes,

facilities,

origins,


destinations,

\color{blue}{10 in total}
& 
Inputs listed include GIS information, attributes related the network such as traffic or disruptive information, location for facilities, origins and destinations.
& 
FTOT
\\

\hline
Describe the geographical components of the FAF dataset? &

\textbf{PREFIX:} rdf: \url{<http://www.w3.org/1999/02/22-rdf-syntax-ns#>}

\textbf{PREFIX:} rdfs: \url{<http://www.w3.org/2000/01/rdf-schema#>}

\textbf{PREFIX:} owl: \url{<http://www.w3.org/2002/07/owl#>}

SELECT ?class ?individual
WHERE \{
    
    ?class rdfs:subClassOf \url{<<---URL--->/ontology#Geography>} .
    ?individual rdf:type ?class .
\}

& 
DomesticOrigin,
ForeignOrigin,
DomesticDestination,
ForeignDestination,

\color{blue}{9 in total}
& 
Geographical components, such as origin and destination.
& 
FAF
\\

\hline
List the main transportation hubs? &

\textbf{PREFIX:} rdf: \url{<http://www.w3.org/1999/02/22-rdf-syntax-ns#>}

\textbf{PREFIX:} rdfs: \url{<http://www.w3.org/2000/01/rdf-schema#>}

\textbf{PREFIX:} owl: \url{<http://www.w3.org/2002/07/owl#>}

SELECT ?region
WHERE {
  ?region rdf:type ontology:Regions .
}



& 
Mobile-Daphne-Fairhope,
Orlando-Deltona-Daytona-Beach,
Chicago-Naperville,
Tucson-Nogales,
St-Louis-St-Charles-Farmington,
Los-Angeles-Long-Beach,
Philadelphia-Reading-Camden,

\color{blue}{24 in total}
& 
List of individual hubs identified by the SPARQL query on the regions class within the FAF ontology.
& 
FAF
\\

\hline
\end{tabular}
\end{table}

\subsection{\rev{Decision Support Strategies - Formulation}\label{sec:DSWF}} 
\begin{figure}[!htp]
    \centering
    \includegraphics[width=\textwidth]{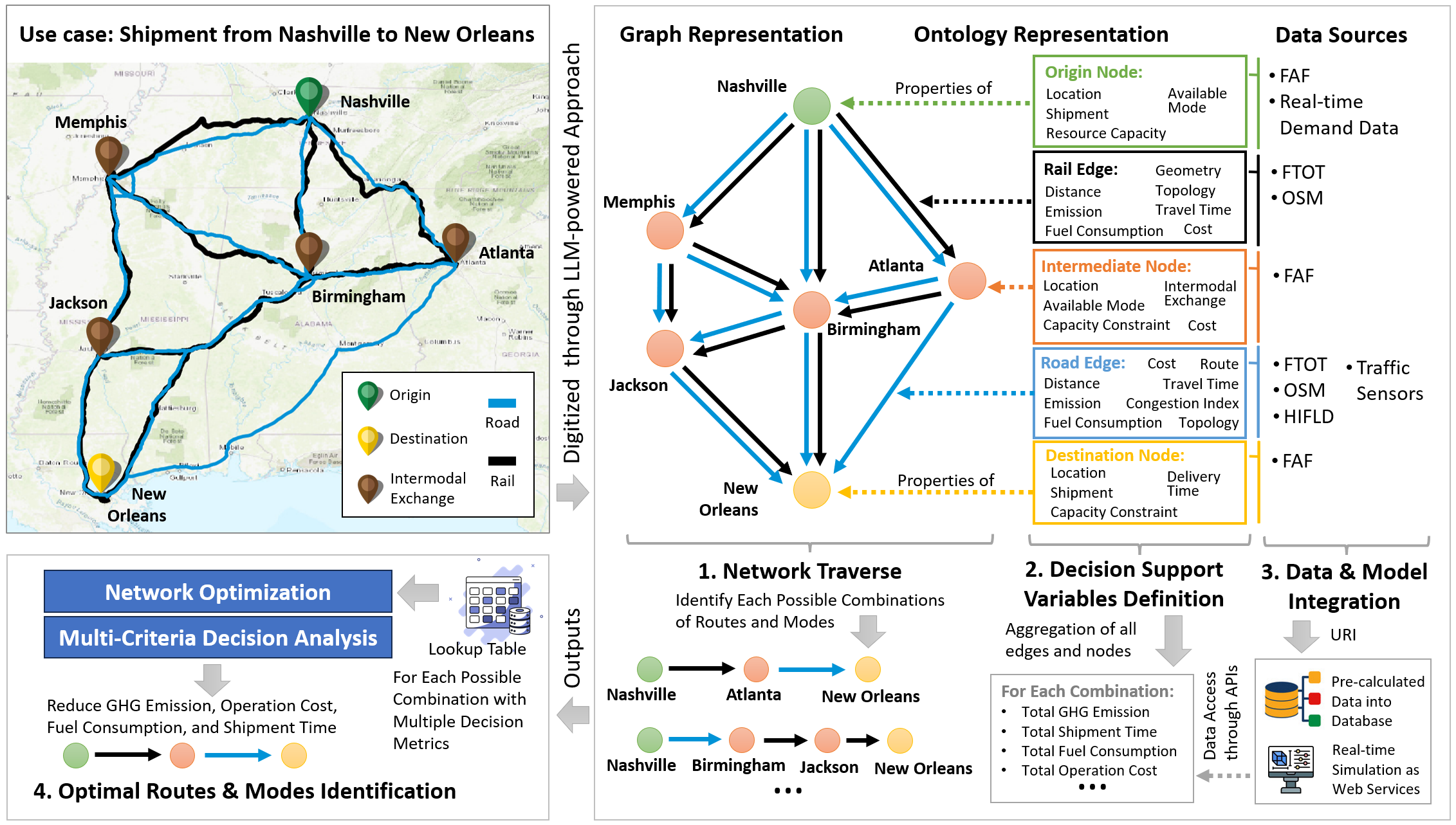}
    \caption{\rev{Conceptual} decision support workflow that is driven by the scenario-based ontology created using the LLM-powered autonomous workflow. }\label{fig:ontology_drive_routes}    
\end{figure}

\revv{Our ontology-driven decision support workflow aims to optimize urban logistics systems across large geographic areas, showcasing the versatility and effectiveness of the LLM-powered approach. By defining CQs aligned with the optimization goals, SPARQL queries are crafted to navigate the AI-generated knowledge representation. This process aids in problem formulation and the selection of decision support metrics, as illustrated in Table \ref{tab:FAF_ontologyUsecause}. These queries return specific concepts, variable names, and relevant metadata, providing urban researchers with a clearer understanding of the available data and simulation resources. The same approach can also be used to identify constraints from the ontology, offering an intuitive and interactive method for identifying pertinent variables from integrated data systems, thereby supporting the development of effective decision support workflows.} The case study presented in this section utilizes a subset of FAF and FTOT data to establish a use case for identifying the optimal combination of shipping modes and routes from Nashville to New Orleans. The conceptual workflow is depicted in Figure \ref{fig:ontology_drive_routes}. Traditional manual approaches to decision optimization often require extensive domain expertise in the dynamics of intermodal freight transportation events, along with data- and simulation-driven insights for quantifying various decision support criteria, such as operational costs and emissions.

By employing our prototyping methodology for creating scenario-based ontology, we have successfully developed a digital representation of knowledge sourced from the technical manuals of major intermodal freight transportation datasets and simulations. This digital knowledge representation is exported into an OWL format, providing an ontology-driven pipeline to outline the decision support workflow and structure data and simulation outputs within the domain context. Our ontology consists of three major components, as the following: 
\begin{itemize}
    \item[a)] Graph Representation is a graph data model that captures the topological properties (connectivity) of the freight transportation network. It is derived by the LLM using information from the FAF dataset manual. The model represents various modes and routes of freight transportation between key \rev{OD} pairs, which are represented by US cities that are major logistic and transportation hubs. This graph is primarily used to navigate between any user-defined origin-destination pairs. 
    \item[b)]  Ontology Representation delineates the major entities, along with their relationships and properties, within the intermodal freight transportation network as represented by the graph model. This ontology is derived from the information outlined in the manuals of both the FAF and FTOT datasets. It provides a foundational data structure that maps data variables—used to characterize the physical dynamics and performance of intermodal transportation processes—to their corresponding graph entities. 
    \item[c)] Data Sources comprise all the necessary multi-domain datasets required to provide data- and simulation-driven insights, complete with essential decision variables. These insights aid in making informed decisions aimed at optimizing operational costs and delivery times while also reducing fuel consumption and GHG emissions. These data sources are organized through a scenario-based ontology and are mapped to corresponding network entities (nodes and edges). Subsequently, they are ingested into a relational database, utilizing a data schema derived from the ontology to organize the data to enable rapid data retrieval for decision support purposes.
\end{itemize}

Leveraging these ontological components, this pipeline facilitates the data warehousing, management, and decision workflow formulation during the development of a data- and simulation-driven decision support system for intermodal freight transportation optimization. The pipeline entails four steps:
\begin{itemize} 
    \item [a)] The ``Network Traverse" refers to the systematic process of navigating through a transportation network to identify all possible combinations of routes and modes between an OD pair. This process involves assessing various transportation modes—such as road, rail, and river—and their interconnections, along with transit locations represented as intermediate nodes within the FTOT network. Each mode and route combination is characterized by a series of nodes that signify the origin, destination, and intermodal transits, connected by edges that represent the transitions between modes. This comprehensive approach ensures a thorough exploration of the network, facilitating optimal route and mode selection.
    \item [b)] The ``Decision Support Variable Definition" is a critical process that identifies key variables essential for characterizing the physical processes involved in intermodal transportation. These variables are vital for determining the most efficient, cost-effective, or fastest combinations of routes and modes available. During the ontology creation stage with LLM, structured prompts are developed. These prompts guide the ChatGPT API in extracting relevant variables from the technical manuals of the FAF and FTOT datasets, focusing on the efficiency and sustainability aspects of the freight transportation system. This structured approach ensures that all critical factors are considered in decision-making processes, enhancing the system's overall performance and sustainability.
    \item[c)] The ``Data and Model Integration" is facilitated by encoding the Uniform Resource Identifier (URI) of specific data or simulations within a relational database, accessible via an API or hosted as a web service, into the ontology as properties of the graph entities. This method ensures that the freight transportation data and simulations are readily available during the network traversal. As the network is traversed to identify an individual combination of route and mode, decision variables at each graph entity are aggregated into decision metrics for the entire route combination, such as total greenhouse gas (GHG) emissions and total operational costs. This integrated setup enhances the efficiency of accessing and utilizing data, thereby optimizing the decision-making process in freight transportation management. The aggregated decision metrics for each route and mode combination are consolidated into a relational lookup table, which serves as a foundation for further decision analysis.     
    \item[d)] The ``Optimal Mode and Route Identification" process utilizes optimization algorithms on a lookup table filled with decision metrics to systematically evaluate and select the most effective transportation options. This evaluation considers multiple, potentially conflicting criteria such as GHG emissions, operational costs, fuel consumption, and shipment time. By integrating both network optimization and Multi-Criteria Decision Analysis, this approach ensures a comprehensive and reliable determination of the optimal route and mode, balancing various transportation needs and environmental impacts effectively.
\end{itemize}

Our methodology leverages the capabilities of LLMs to automate the digitization of domain knowledge encapsulated in text and figures within technical manuals, urban datasets, simulations, and research articles. This knowledge is transformed into interoperable ontology formats, which serve as the foundation for developing data models. These models structure essential data and simulation outputs to support decision-making processes. The real-time generated lookup table from the ontology-driven data model is a critical component. Together, these elements form essential parts of urban decision support systems or digital twin applications, enhancing the efficiency and effectiveness of urban planning and management.

\begin{figure}[!htp]
    \centering
    \includegraphics[width=\textwidth]{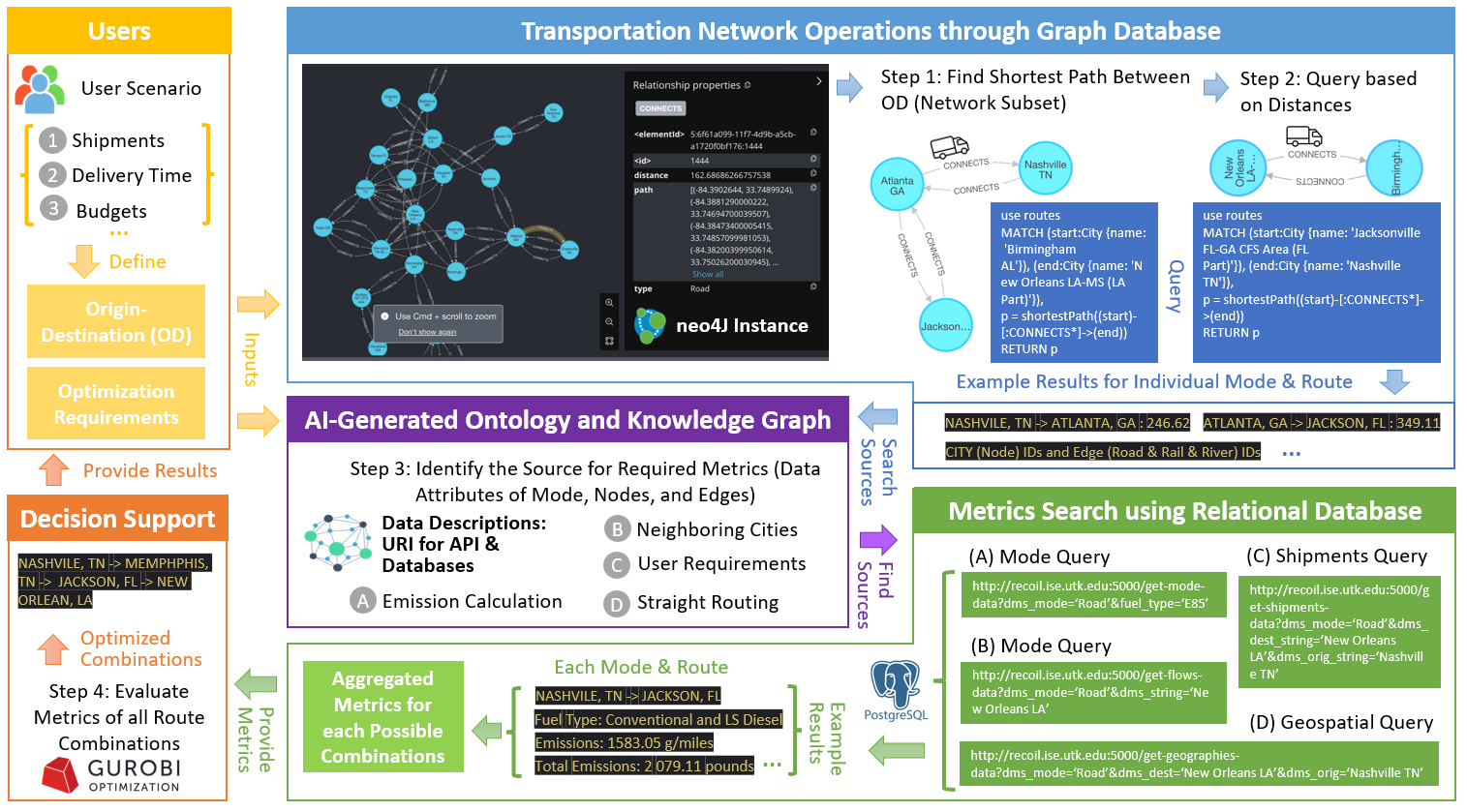}
    \caption{  \rev{Technical implementation of the decision support workflow utilizes the proposed ontology-driven methods to guide the design and queries of PostgreSQL and neo4j databases.}}\label{fig:optimization}    
\end{figure}

\rev{\subsection{Decision Support Strategies - Implementation} }
\rev{
We implemented our proposed methodology using the AI-generated ontology and its knowledge graph to provide data source descriptions, guiding the design and queries of databases that are essential components of the data provisioning system. Figure \ref{fig:optimization} demonstrates the technical pipeline, outlining the processes involved in supporting decision-making to reduce GHG emissions, operational costs, fuel consumption, and shipment time using structured data and simulation outputs.
The first step in the technical pipeline is defining OD pairs and optimization requirements by the target users. These inputs are directly fed into the neo4j graph database and the AI-generated ontology and knowledge graph to configure the optimization scenario. Based on the OD, the graph database conducts a shortest path query that identifies a list of possible combinations of modes and routes between the OD based on transport distance. Example queries of the graph database operations are provided in Figure \ref{fig:optimization}. Through the query, each combination of mode and route will be identified, specifying the mode with fuel types, node IDs representing hubs and intermodal transit points, and edge IDs describing the attributes of the transportation path (e.g., rail, road, and river). }

\rev{By configuring the ontology, we define the constraints and requirements of user scenarios, guiding the inclusion and exclusion of specific decision criteria. In the demonstration, we showcase a scenario aimed at identifying the most optimized shipping solution between Nashville and New Orleans, as conceptually depicted in Figure \ref{fig:ontology_drive_routes}. This scenario will be optimized based on criteria including GHG emissions, shipment distance, and shipment time. 
Based on the query results from the graph database, the AI-generated ontology defines the decision support metrics and provides data and simulation descriptions through its knowledge graph. This guides the discovery of essential decision metrics stored in the PostgreSQL relational database. This setup allows for the future inclusion of more ontology-driven parameters, such as route disruptions, free-speed travel, and capacity. For this test scenario, we focus solely on distance, node vicinity, metrics, and constraints. 
}

\rev{Once the variables characterizing decision support metrics are retrieved for each combination of modes and routes, a Python script aggregates these metrics to calculate the total amount of GHG emissions, time, and cost. The various combinations identified through the graph database, along with their defined metrics, are then used as inputs for the GUROBI optimization engine.}\\

\section{Limitation and Future Work}
Throughout the course of development and evaluation, we have identified two major limitations associated with our current AI-powered methodology for generating ontologies. These limitations, along with their potential solutions through future work, are outlined as follows:

\begin{enumerate}
\item Without precisely engineered prompts, the ChatGPT API may exhibit sensitivity, leading to subtle variability and inconsistent results across different iterations. The robustness of the model can be enhanced by employing more advanced techniques, such as fine-tuning. These methods adapt pre-trained LLMs to generate ontologies using specific datasets or under particular conditions, thereby minimizing potential variability in the results from different model runs.
\item There is a limitation on the maximum number of tokens that the ChatGPT API can process per request. This constraint can be circumvented by partitioning, optimizing, and refining the input data, combined with advancements in LLM research.
\item \rev{Potential hallucinations of LLMs could jeopardize the accuracy and reliability of AI-generated ontologies and knowledge graphs. To minimize these hallucinations, previous studies have proposed matching techniques that utilize vocabularies defined in existing domain ontologies to validate AI-generated content \citep{caufield2024structured}. Additionally, leveraging external knowledge bases through the recent emergence of RAG can also help reduce hallucinations. Our method demonstrates a qualitative analysis that logically addresses the competency questions proposed through SPARQL queries.}

\end{enumerate}
We aim to address these limitations through our future work \rev{by deploying and fine-tuning a local pre-trained model using Ollama.}

\section{Conclusion} \label{sec:Discussion}
In this study, we introduced an AI-powered methodology designed to automate the generation of scenario-based scientific ontologies by leveraging the text understanding and reasoning capabilities of LLMs. This methodology was implemented in a prototype autonomous agent using Python NLP libraries and the ChatGPT API. Our primary goal is to support the development of urban decision support systems by automating time-consuming and labor-intensive tasks that usually require extensive domain knowledge. Our automated approach facilitates the integration of data resources and simulations, thereby enhancing informed decision-making.

\revv{We tested our methodology by generating a partial segment of the Pizza Ontology} and compared the AI-generated outputs with the expert-created Pizza Ontology, validating the reliability and feasibility of our approach. Through a case study, we demonstrated how our methodology could be applied to real-world urban data, such as the FAF and FTOT datasets, to aid in the development of an ontology-driven urban decision support system. Although still in its early stages, our approach showcases the potential and feasibility of using an autonomous workflow to generate ontologies that support scientific software and database development, as well as guide the formation of decision support strategies. This initiative highlights the next generation of decision support systems and digital twins that leverage large-scale urban data to optimize complex urban systems.

With significant promise for achieving autonomous decision support systems, we believe our methodology can be further extended and applied to creating various scenario-based ontologies. This extension could lower technical barriers and enhance the ability to address complex urban system challenges. Future work will focus on applying fine-tuning techniques to instruct LLMs on generating more specialized ontologies, as well as developing software modules and codes that serve as critical components of an urban decision support system.  

  
\section{Acknowledgements} \label{sec:Acknowledgements}
This work was supported in part by the U.S. Department of Energy's Advanced Research Projects Agency-Energy (ARPA-E) under the project (\#DE-AR0001780) titled \textit{``A Cognitive Freight Transportation Digital Twin for Resiliency and Emission Control Through Optimizing Intermodal Logistics"} (RECOIL).

\small


\end{document}